\documentclass[journal]{IEEEtran}
 
\ifCLASSINFOpdf
\else
\fi
\usepackage{soul}
\usepackage{xcolor} 
\soulregister\cite7
\soulregister\citep7 
\soulregister\citet7  
\soulregister\ref7  
\soulregister\pageref7

\usepackage{amsfonts}
\usepackage{graphicx}
\usepackage{amsmath,amssymb} 

\usepackage{algorithm}
 
\usepackage{algorithmic}
\usepackage{epsfig}

\usepackage{graphicx}

\usepackage{here}
\usepackage[pagebackref=true,breaklinks=true,letterpaper=true,colorlinks,bookmarks=false]{hyperref}
\usepackage{caption}
\begin{document}
%
\title{Subtype-aware Dynamic Unsupervised \\ Domain Adaptation}
%
%
%

\author{Xiaofeng~Liu,
        Fangxu~Xing,
        Jane~You,
        Jun~Lu,
        C.-C.~Jay~Kuo,~\IEEEmembership{Fellow,~IEEE,}
        Georges~El~Fakhri,~\IEEEmembership{Fellow,~IEEE,}~Jonghye~Woo,~\IEEEmembership{Senior Member,~IEEE}
\thanks{X. Liu, F. Xing, G. El Fakhri, and J. Woo are with the Gordon Center for Medical Imaging, Department of Radiology, Massachusetts General Hospital and Harvard Medical School, Boston, MA, 02114, USA.\\~~Jun Lu is with the Department of Neurology, Beth Israel Deaconess Medical Center and Harvard Medical School, Boston, MA, USA.\\~~Jane You is with the Department of Computing, The Hong Kong Polytechnic University, Hong Kong.\\~~C.-C. J. Kuo is with Ming Hsieh Department of Electrical and Computer Engineering, University of Southern California, Los Angeles, CA, 90007, USA.}

\thanks{Manuscript received on Jan 15, 2021; revised on June 25, 2021 and Dec 14, 2021, accepted on June 26 2022.}}

%
%

\markboth{Published on IEEE Trans on NNLS,~June~2022}%
{Shell \MakeLowercase{\textit{et al.}}: Manuscript Submitted to IEEE Trans on}
%



\maketitle
 
\begin{abstract}

Unsupervised domain adaptation (UDA) has been successfully applied to transfer knowledge from a labeled source domain to target domains without their labels. Recently introduced transferable prototypical networks (TPN) further addresses class-wise conditional alignment. In TPN, while the closeness of class centers between source and target domains is explicitly enforced in a latent space, the underlying fine-grained subtype structure and the cross-domain within-class compactness have not been fully investigated. To counter this, we propose a new approach to adaptively perform a fine-grained subtype-aware alignment to improve performance in the target domain without the subtype label in both domains. The insight of our approach is that the unlabeled subtypes in a class have the local proximity within a subtype, while exhibiting disparate characteristics, because of different conditional and label shifts. Specifically, we propose to simultaneously enforce subtype-wise compactness and class-wise separation, by utilizing intermediate pseudo-labels. In addition, we systematically investigate various scenarios with and without prior knowledge of subtype numbers, and propose to exploit the underlying subtype structure. Furthermore, a dynamic queue framework is developed to evolve the subtype cluster centroids steadily using an alternative processing scheme. Experimental results, carried out with multi-view congenital heart disease data and VisDA and DomainNet, show the effectiveness and validity of our subtype-aware UDA, compared with state-of-the-art UDA methods.
\end{abstract}

 
\IEEEpeerreviewmaketitle

\section{Introduction}
 
Recent advances in deep learning (DL) have shown promising performance in many supervised learning tasks \cite{che2019deep}. In supervised learning, large-scale labeled training data are the key requirement for accurate model fitting at the training stage \cite{zou2019confidence}. Although sufficient labels are available for some applications, such as ImageNet, acquiring large-scale labeled manual data is still challenging in many applications, including medical imaging, partly because of expert-driven annotation costs \cite{liu2018ordinal}. Besides, conventional empirical risk minimization tasks rely on independently and identically distributed ($i.i.d.$) training and testing data \cite{che2019deep,liu2019feature}. In many real-world tasks, however, a variety of disparate data distributions, i.e., domain shift, between different domains exist, which hinders the application of supervised learning on new domains \cite{kouw2018introduction,liu2021recursively}. For instance, in medical imaging, domain shift arises due to differences in scanners, protocols, and modalities, which also causes the high cost of collecting large-scale labeled training data annotated by clinicians \cite{shen2017deep}.

Unsupervised domain adaptation (UDA) is an emerging technique, aiming at transferring domain knowledge learned from a labeled source domain to target domains with the help of unlabeled target data \cite{zou2019confidence,shao2014transfer,zhang2019guide}. To counter the discrepancy in data distributions between two domains, the class-wise conditional alignment w.r.t. $p(x|y)$ is typically employed \cite{pan2019transferrable,zhang2013domain,long2018conditional}. The widely used UDA methods make use of adversarial learning and maximum mean discrepancy (MMD) to align the latent space distribution of sample data $x$, i.e., $p(f(x))$, where $f(\cdot)$ is a feature extractor. Using the Bayes' theorem $p(f(x)|y)=\frac{p(y|f(x))p(f(x))}{p(y)}$ and assuming no concept shift w.r.t. $p(y|f(x))$ and label shift w.r.t. $p(y)$, the alignment of conditional distribution $p(f(x)|y)$ can be achieved by matching $p(f(x))$. Another challenge in UDA is the difference in the proportion of each class between two domains, thus leading to the label shift w.r.t. $p(y)$ \cite{zhao2019learning,yang2019hybrid}.

\begin{figure}[t]
\centering 
\includegraphics[width=9cm]{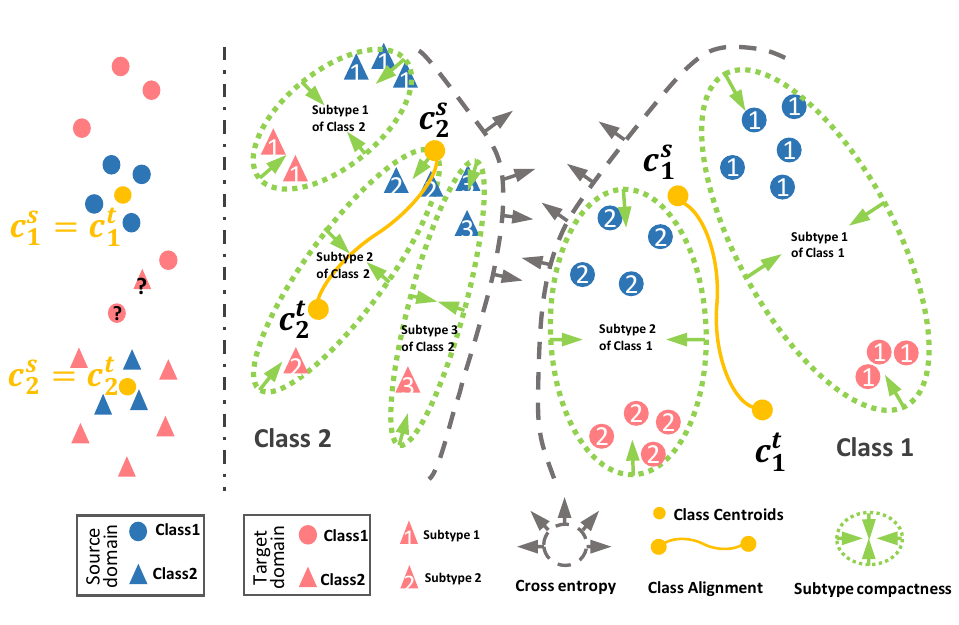}\\
\caption{An example of a failure case of prototypical UDA (left) and illustration of our proposed subtype-aware UDA (right). Note that the class label in the target domain and the subtype label in both source and target domains are inaccessible.}\label{fig:11} 
\end{figure}

\begin{figure}[t]
\begin{center}
\includegraphics[width=0.9\linewidth]{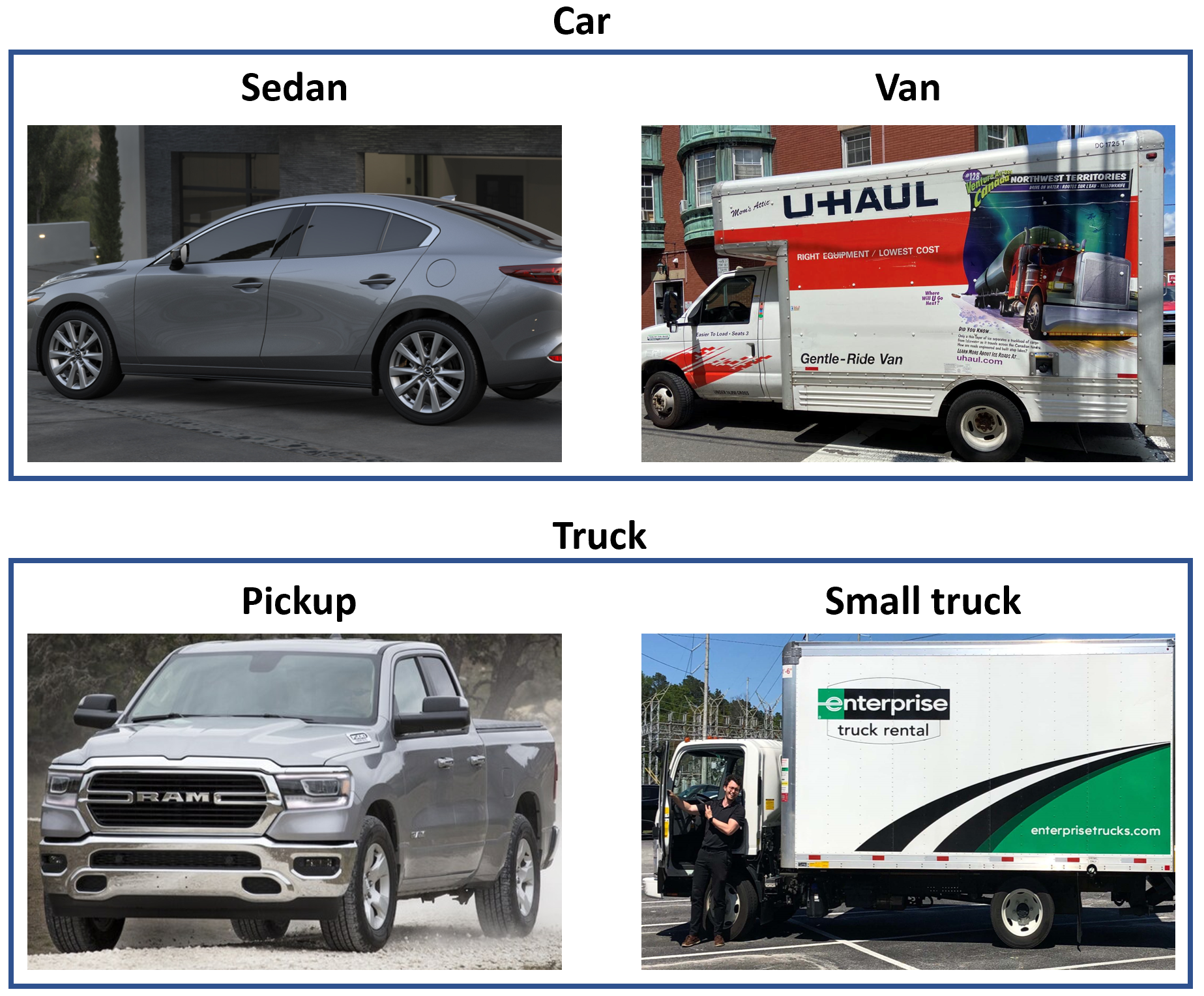}
\end{center} 
\caption{Target examples of the car and truck classes from the VisDA17 real-world image set. Noticing that the subtype labels are not offered.}
\label{fig:sp1}
\end{figure}

Recently, transferable prototypical networks (TPN) \cite{pan2019transferrable,jing2020adaptively} is proposed to promote the class separation in the source domain using a cross-entropy (CE) loss, by minimizing the distance of the class centers between two domains for the class-level conditional alignment w.r.t. $p(x|y)$. The CE loss used in TPN, however, cannot minimize the inner-class variation in the source domain \cite{liu2016large,wen2016discriminative}. Additionally, the optimization objective of the centroid closeness cannot guarantee the inter-class separation in the target domain. This problem is well illustrated in Fig.~\ref{fig:11} left, where while the class centroids are perfectly aligned, there is a possibility that the sparsely distributed target samples are easily misclassified. One way to solve this would be by simply encouraging the cross-domain inner-class feature distribution compactness. In many real-world applications, however, the unlabeled subtypes in a class can be highly diverse, and thus an underlying local distribution forms an ensemble distribution. For example, the VisDA17 dataset \cite{visda2017} has the car, truck, and bus classes. The car class contains the sedan, SUV, van, etc., which have large inner-class variations, and some of its subtypes can be visually similar to the bus and truck classes, as shown in Fig. \ref{fig:sp1}. The shared pattern among the sedan, SUV, and van may not be exclusive for the bus and truck classes. Likewise, the subtypes of a disease can differ significantly with one another \cite{yeoh2002classification}. The shared characteristics among two subtypes may not be exclusive for class-wise discrimination. In such cases, it would be more reasonable to explore the subtype-wise patterns. Although the fine-grained label can be useful for the coarse meta-class prediction \cite{chen2019understanding}, it is usually costly or infeasible to label the fine-grained subtype for a large scale dataset \cite{yeoh2002classification,sonpatki2020recursive}; however, the manner in which the underlying subtype, without the subtype label, is exploited in both source and target domains for a variety of tasks poses a challenge.


The problem of conditional and label shifts w.r.t. subtypes, on the fine-grained subtype level, can also arise, in which we do not have access to the subtype label in either source or target domain. For example, we usually have the label of the pedestrian only, without the gender label, although the cross-gender difference can be significant. In addition, from winter to summer, the male/female of the pedestrian class can have different dress changes (shorts/skirt), which results in \textit{subtype conditional shift}. Furthermore, the incidence of disease subtypes or the ownership rate of the sedan/SUV is usually varied across areas, which results in \textit{subtype label shift}. The proportion difference on the subtype level can generally be more significant than the class level \cite{wu2019domain}. In the implementation, some of the subtypes may be undersampled with the limited batch size. These observations motivate us (1) to expand the definition of class conditional and label shifts \cite{kouw2018introduction} to the fine-grained subtype level, i.e., $p(f(x)|k)$, and (2) to re-formulate that $p(k)$ could differ across domains for subtype $k$. As such, a more practical assumption would be the conditional and label shifts on both the class and subtype levels.

To address the aforementioned challenges, we propose to adaptively achieve both class-wise separation and cross-domain subtype-wise compactness. Specifically, we resort to feature space metric learning with intermediate pseudo-labels. In our subtype-aware UDA, we do not have access to the class label in the target domain and the subtype label in both the source and target domains. An online clustering scheme is proposed to explore the underlying subtype structure in an unsupervised manner. With the prior knowledge of subtype numbers, a straightforward choice would be the widely used $K$-means clustering, which defines $k$ to be the subtype numbers. Due to taxonomies present in the data, however, it is challenging to identify the exact subtype number. We thereby further extend our method to the case of uncertain subtype numbers, by using the reliability-path to capture the underlying subtype structure with an adaptive sub-graph scheme. The sub-graph scheme can be scalable over several classes and subtypes, with a few meta hyperparameters shared between clusters. The involved samples of subtypes are required to be adjacent to their subtype centroid in the feature space to directly impose the subtype-wise distribution compactness. {The presence of the subtype is a general problem for any kind of DA method. In our method, assigning a pseudo-subtype label in each iteration requires the pseudo-class label to be assigned first. Then, the subtypes within each class are explored. Notably, the pseudo-class label assignment is a typical operation in the prototype-based methods. Thus, the prototype-based methods can be used as our baseline.}

In addition, given the possible unstable learning caused by the undersampling of subtypes and the subtype label shift, we base our training on a dynamic memory framework to steadily evolve the cluster centroids to keep the network stably updated. Specifically, we design and maintain feature queue and centroid memory modules to store the sample features of several batches and the current centroids. These two modules are updated alternatively, and the features are refined with a momentum scheme, thereby allowing more representative sampling and clustering with low space cost. We breakdown the abrupt off-line entire dataset clustering \cite{caron2018deep} into two steps: (1) steady memory updates and (2) per batch pseudo label re-assigning. Moreover, our sub-graph and subtype centroid constructions are robust against the undersampling or label shift of any subtype.
 
This work extends our previous work \cite{Liu2021subtype} in the following significant ways:
 
\textbf{$\bullet$} An end-to-end dynamic queue framework with steady memory updates and per batch pseudo label re-assigning, which renders our subtype-aware UDA to be a practical and scalable solution.

\textbf{$\bullet$} The dynamic queue framework is able to process the imbalanced and undersampled subtypes, by involving the subtype cluster centroids steadily along with the iterations with an alternative processing scheme.
 
\textbf{$\bullet$} We demonstrate the generality of our subtype-aware UDA scheme in general large-scale UDA benchmarks, i.e., VisDA17, DomainNet, and GTA5toCityscapes. 

\textbf{$\bullet$} Moreover, we carry out all experiments using a novel dynamic queue framework and provide a systematical subtype analysis with the two-level label structure of DomainNet.
 
The main contributions of this paper are summarized as follows:

\textbf{$\bullet$} We propose to adaptively explore the subtype-wise conditional and label shifts in UDA without the subtype labels, and explicitly enforce the subtype-aware compactness.


\textbf{$\bullet$} We systematically investigate cases with or without the prior information of subtype numbers. Our reliability-path-based sub-graph scheme can effectively explore the underlying subtype local distributions with only a few meta hyperparameters in an online fashion.

\textbf{$\bullet$} Our alternatively updated dynamic memory queue and centroid module can be a practical and scalable solution to process the imbalanced and undersampled subtypes. It is possible to steadily evolve the cluster centroids to keep the network stably updated.

Taken together, this is the first attempt at exploring the underlying subtype structure, which does not have a subtype label in both the source and target domains. We evaluate the effectiveness of our framework on multiple benchmarks with different backbone models.

\section{Related Work}

Recent advances in DL have achieved tremendous milestones in both computer vision and medical imaging areas \cite{li2019relaxed}. DL has drastically converted the manner in which various features are extracted and fed into a prediction model into jointly learning both features and the prediction model from raw data in an end-to-end fashion~\cite{liu2020severity,liu2020auto3d,liu2019unimodala,liu2019unimodalb}. In order for DL to be successfully applied to a variety of applications, massive annotated samples are typically demanded, where an $i.i.d.$ assumption of training and testing data is required to ensure a seamless application of the trained model \cite{che2019deep,liu2020energy}. 

A few challenges arise in applying DL to real-world applications. For example, a DL model that is trained on an annotated source domain does not generalize well on different target domains, because of the domain shift \cite{liu2021adapting,liu2021generative,wang2019transferable}. In addition, collecting the large scale annotated training data in a novel environment can be costly or even prohibitive. Therefore, it is required to migrate domain knowledge learned from an annotated source domain with UDA to unseen target domains \cite{8578933,7410820}.

Based on the shift content, domain shifts \cite{kouw2018introduction,zhang2019bridging} can be categorized into four groups, including (1) covariate shift w.r.t. $p(x)$, (2) conditional shift w.r.t. $p(x|y)$, (3) label shift w.r.t. $p(y)$, and (4) concept shift w.r.t. $p(y|x)$. To date, each shift has been studied independently, with an assumption that the other shift conditions are invariant across domains \cite{kouw2018introduction}. In many applications, the conditional shift \cite{magliacane2018domain,li2018domain} is more likely to occur than the covariate shift \cite{kouw2018introduction}. The label shift \cite{chan2005word} (a.k.a. the target shift) occurs when the proportion of each class is different between source and target domains. We note that the partial UDA can also be considered a label shift \cite{li2020deep}. The concept shift \cite{kouw2018introduction} arises when classifying a tomato into a vegetable or fruit in different countries; however, the concept shift is usually not a common problem in popular object classification or semantic segmentation tasks. The conditional and target shifts are investigated in a casual system ($y\rightarrow{x}$), with a simple linearity assumption \cite{zhang2013domain,gong2016domain}. However, these approaches could be sensitive to a dataset scale.

\begin{figure}[t]
\centering
\includegraphics[width=9cm]{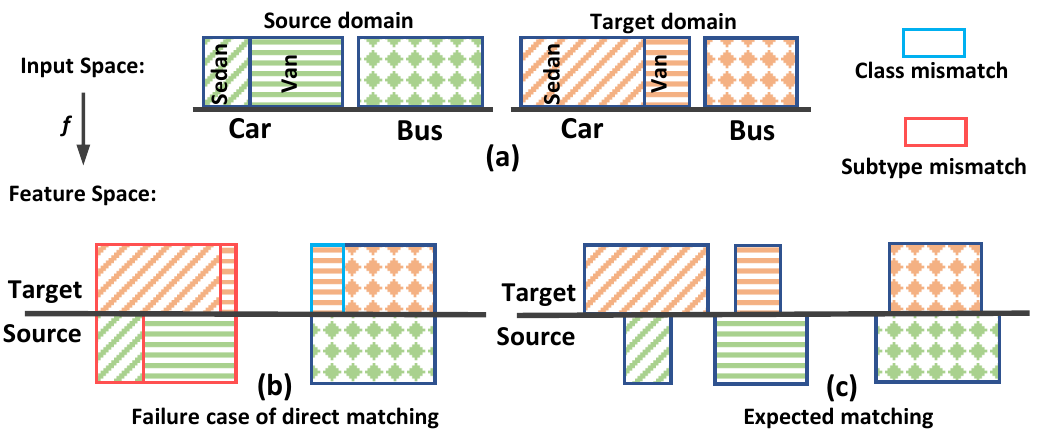}\\ 
\caption{A failure case of class alignment and our expected outcome of subtype alignment under subtype label shift. {In this figure, we considered two classes, including car and bus, and the car class has two subtypes including sedan and van. There are label shifts at both class and subtype levels, i.e., the proportion (the area in the figure) of classes and subtypes in two domains are different.}}\label{fig:22} 
\end{figure}

The objective of traditional UDA is to simply align the covariate shift (i.e., only $p(x)$ shift). Discrepancy-based techniques, such as minimizing MMD, have been widely used to address the covariate shift in source and target domains, by minimizing discrepancies defined on different layers of a shared model between source and target domains \cite{long2015learning}. 

In recent years, adversarial learning makes use of an extra domain discriminator to promote domain confusion \cite{tzeng2017adversarial}. These approaches are based on the assumption that the proportion of the labels remain invariant between source and target domains \cite{moreno2012unifying}. However, the label shift (i.e., only $p(y)$ shift) is common in many real-world applications \cite{kouw2018introduction}. {Recently, \cite{zhang2019bridging} proposes a margin disparity discrepancy, a novel measurement with rigorous generalization bounds. The attention scheme is used in \cite{wang2019transferable} for UDA. Learning the domain invariant and class discriminative features is investigated in \cite{li2018domain}.}

As stated above, in addition, the subtype conditional and label shifts are a realistic assumption, which occurs in many applications, but those shifts pose more challenges, especially when the subtypes are undersampled. As shown in Fig. \ref{fig:22}, we adapt the class label shift concept in \cite{wu2019domain} to the subtype. {According to \cite{wu2019domain}, if the source distribution contains 50\% car and 50\% bus, while the target distribution contains 70\% car and 30\% bus, then successfully aligning these distributions in a representation space requires a classifier to predict the same fraction of car and bus on the source and target data. If one achieves 100\% accuracy on the source data, then target accuracy will be at most 80\%. Considering the van class can be confused with the bus class, it is highly likely that the van class is misclassified as the bus class, as denoted with the blue rectangle in Fig. \ref{fig:22}. Similarly, there also has a mismatch at a subtype level, as denoted with the red rectangles. In Fig. \ref{fig:22} (c), we show the expected subtype-wise alignment.}

Another line of research in UDA is to use pseudo-labels in the target domain \cite{zou2019confidence,liu2020energy}. For example, the use of pseudo-labels for target class centroids calculation is proposed to match source class centroids~\cite{chen2019progressive}. Contrastive adaptation network (CAN) \cite{kang2019contrastive} is proposed to use the pseudo-labels in the target domain for contrastive domain discrepancy estimation. Since the pseudo-labels are likely to be noisy compared with manual labels, a new approach~\cite{gu2020spherical} is presented to measure its correctness by the Gaussian-uniform mixture model. In the present work, we instead propose a noise-robust sub-graph scheme with a simple online semi-hard mining method. 

The ``subdomain" adaptation approach proposed in \cite{zhu2020deep} also focuses on the alignment of class-wise matching, and each subdomain corresponds to a class. Our proposed subtype, however, is more fine-grained than the class for which we do not have the subtype label in either source or target domain.

Based on self-labeling, TPN \cite{pan2019transferrable} is proposed to achieve the class-wise conditional alignment, by matching the source and target class centers \cite{chen2019progressive,jing2020adaptively}. Additionally, it is possible to set source centers as class protocols for classification. While its CE loss can enforce the source class-wise separation \cite{liu2016large}, aligning the centers does not enforce the compactness of both source and target samples, since the target samples can distribute sparsely with considerable within-class variations. In addition, the prototypes in \cite{pan2019transferrable} can be sensitive to the label shift.    

The center loss \cite{wen2016discriminative} has been developed to enforce the compact distributed features for face recognition. Then, new approaches \cite{liu2017adaptive,xu2020reliable} are presented to adapt the center loss to metric learning and optimal transport based methods. Similarly, the large margin softmax, ArcFace, and CosFace losses are proposed to simultaneously enforce inter-class separation and inner-class closeness \cite{liu2016large}. {Recently, \cite{tang2020unsupervised} encourages class-wise closeness across domains. Similarly, \cite{toldo2021unsupervised} proposes to cluster the representation of each pixel. However, those approaches only rely on the class label supervision, and the underlying subtype distributions and shifts are largely ignored. When the discrepancy within a class is large, it is unreasonable to strictly extract the shared pattern.

In addition, in \cite{tang2020unsupervised,toldo2021unsupervised}, the samples in each cluster are known, and we only need to calculate the distance. Instead, without the subtype label in both domains, the unsupervised clustering (e.g., $K$-means or sub-graph) can be more challenging to identify the examples in each cluster.}

Our approach exploring the subtype in an online fashion is closely related to unsupervised clustering \cite{caron2018deep,yu2020semisupervised}. Along this line of research, $K$-means is used as an unsupervised initialization for convolutional layers sequentially, following a bottom-up fashion \cite{coates2012learning}. Note that the fine-grained labeling of a subtype can be costly and challenging, especially in the medical area. As such, several efforts have been made for subtype discovery with unsupervised clustering \cite{yeoh2002classification,sonpatki2020recursive}. A deep clustering approach \cite{caron2018deep} is proposed to train a network with $K$-means clustering, following an end-to-end fashion. Moreover, $K$-means for class-level clustering is proposed in open-set DA~\cite{pan2020exploring}, which is a quite common operation, since classification always requires inner-class compactness. That work, however, does not consider the subtype-level clustering. In the present work, we investigate $K$-means \cite{coates2012learning} with the prior information on subtype numbers. In addition, without the need for the number of subtypes, we propose a scalable sub-graph scheme. Furthermore, the global clustering used in \cite{caron2018deep} is used, by simply applying $K$-means on all of the training data of ImageNet, which, however, can be highly costly. In addition, their training of AlexNet takes about 12 days. To address this, efficient dimension reduction and dynamic memory schemes are proposed to simplify the online clustering. Notably, the dynamic memory architecture in our work can be robust to pseudo-label noise and subtype undersampling.

{Generally, since collecting large-scale labeled data in the target domain poses a challenge, labeling partial and scarce target data can be a practical solution \cite{he2020classification}. In addition, the pseudo-label can be utilized, which provides a good connection between semi-supervised learning and unsupervised domain adaptation \cite{zou2019confidence}. For example, recent work \cite{wang2019domain,wang2020unsupervised} explores the scarce labeled target data in UDA with pseudo-label. In our framework, reliability-path based sub-graph construction can be facilitated with some known samples. In the present work, we only consider the case when the number of categories between source and target domains is consistent. Recently, \cite{jing2020adaptively} and \cite{li2020deep} propose to achieve partial UDA with adaptively-accumulated knowledge transfer and a deep residual correction network, respectively. \cite{jin2020minimum} proposes a flexible interface between the closed and partial set UDA. The open set UDA has been proposed to explore the case of non-overlapping class \cite{liu2019separate,fu2020learning}. These works are orthogonal to our contributions to explore the subtype structure. Exploring the subtype structure in semi-supervised UDA or open set UDA can be a promising future direction.}

\section{Methodology} 

In the scenario of UDA, there are a labeled source domain $p^s(x,y)$ and a different target domain $p^t(x,y)$, and the learning system has access to a labeled source set $\{(x_i^s,y_i^s)\}$ drawn from $p^s(x,y)$ and an unlabeled target set $\{(x_i^t)\}$ drawn from $p^t(x,y)$. The class label space of $y_i\in\left\{1,2,\cdots,N\right\}$ is shared for both domains. For indexing $N$ classes, we use $n$. We assume that there are $K_n$ underlying subtypes indexed with $k\in\{1,2,\cdots,K_n\}$ in class $n$, and we do not have access to the fine-grained subtype label in both domains. UDA attempts to learn a discriminative model in the target domain $p^t(x,y)$ following the theorem: \\~

\noindent\textbf{Theorem 1} For a hypothesis $h$ drawn from $\mathcal{H}$, $\epsilon^t(h)\leq$ $\epsilon^s(h)+\frac{1}{2}d_{\mathcal{H}\triangle\mathcal{H}}\{s,t\}+^{\rm{min}}_{h\in\mathcal{H}}[\epsilon^s({h},l_s)+\epsilon^t({h},l_t)]$.\\~

\noindent Here, $\epsilon^s(h)$ and $\epsilon^t(h)$ denote the expected losses with hypothesis $h$ in the source and target domains, respectively. The right side can be the upper bound of the target loss. Since the disagreement between the labeling functions $l_s$ and $l_t$, i.e., $^{\rm{min}}_{h\in\mathcal{H}}[\epsilon^s({h},l_s)+\epsilon^t({h},l_t)]$, can be small, by optimizing $h$ with the source data \cite{ben2007analysis}, UDA aims to minimize the cross-domain divergence $d_{\mathcal{H}\triangle\mathcal{H}}\{s,t\}$ in the feature space of $f(x_i^s)$ and $f(x_i^t)$.

Instead of matching $p^s(f(x))$ and $p^t(f(x))$ \cite{kouw2018introduction}, the prototypical networks attempt to align the class centroids \cite{pan2019transferrable}. However, it poses a challenge to define the decision boundary of the low-density distributed target sample and thus the inherent subtype structure is not fully investigated.
 
In this work, we aim to achieve the class separation in the target domain on the embedding space, when the source domain classes are well-separated, and compactness of the class-wise source-target distribution is encouraged. Additionally, under the fine-grained subtype local structures and their conditional and label shifts, the subtype-wise compact clustering can be a good alternative to achieve the class-wise alignment and compactness. Therefore, we have the following proposition:\\~

\noindent\textbf{Proposition 1.} The class-level compactness can be a particular case of subtype-wise compactness by assigning the subtype number of this class to 1. \\

To tackle the conditional and label shifts at both class-level ($p^s(f(x)|y)\neq p^t(f(x)|y), p^s(y)\neq p^t(y))$ and subtype-level ($p^s(f(x)|k)\neq p^t(f(x)|k), p^s(k)\neq p^t(k))$, we aim to develop a subtype-aware alignment framework with a dynamic clustering scheme as illustrated in Fig.~\ref{fig:11}. 



We organize the method section as follows: 

We first describe our approach to tackle the class-wise separation and alignment in Sec. III.A. Then, we explore subtype-wise compactness for the following scenarios: 1) If we know the subtype number, we use the subtype-aware alignment with $K_n$ prior in Sec. III.B; 2) If we do not know the subtype number, we use the reliability-path based sub-graph construction in Sec. III.C. To speed up the clustering (either $K$-means or sub-graph), we propose queued dynamic clustering in Sec. III.D. Sec. III.E, i.e., optimization and implementation, provides detailed processing. 

\begin{figure}[t]
\centering
\includegraphics[width=9cm]{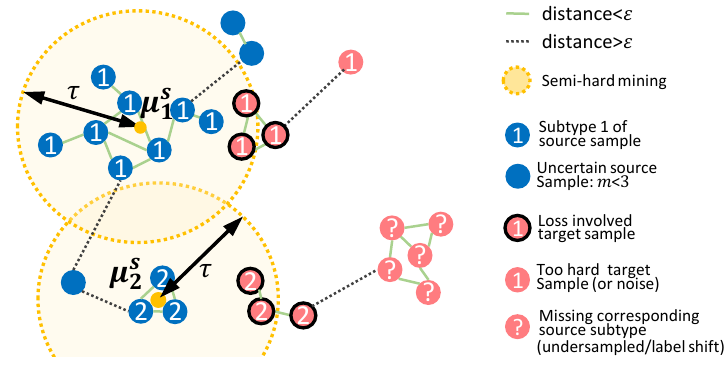}\\ 
\caption{Illustration of reliability-path based sub-graph construction and alignment with $m=3$. \protect\includegraphics[scale=0.3]{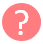} will be assigned to subtype 2 and then be rejected by $\tau$.}\label{fig:33} 
\end{figure}

Our framework has two promising properties. The adaptive property is achieved by the section III.B ``Reliability-path based sub-graph construction" (see Fig. \ref{fig:33}). If we do not know how many subtypes we have, it is possible to cluster the samples adaptively. The dynamic property is achieved by the section III.C ``queued dynamic clustering" (see Fig. \ref{fig:44}), which largely alleviates the memory cost and evolves the centroids smoothly.

\subsection{Class-wise source separation and matching} 

Forcing the separation of classes in the source domain $p^s(x,y)$, i.e., minimizing $\epsilon^s(h)$, can be achieved by the conventional CE loss \cite{liu2016large}. We carry out the classification with the extracted features through a remold of the distance to each class cluster centroid \cite{chen2019progressive,pan2019transferrable}.

For the labeled source data $\{(x_i^s,y_i^s)\}$, we represent the feature distribution of class $n$ with a class centroid \begin{equation}
\begin{aligned}
c_n^s=\frac{1}{M_c^s}\sum_{i=1}^{M_c^s}f(x_i^s), 
\end{aligned}\end{equation}where ${M_c^s}$ is the number of source samples involved. We can directly create a probability histogram for an input source sample $x_i^s$ using the softmax normalized distance between $x_i^s$ and the centroids $c_n^s$. Specifically, it is possible to formulate the likelihood of $x_i^s$ belonging to class $n$ as\begin{equation}
\begin{aligned}
p(y_i^s=n|x_i^s)=\frac{e^{-||f(x_i^s)-c_n^s||_2^2}}{\sum_{n=1}^N e^{-||f(x_i^s)-c_n^s||_2^2}}.\label{eq:1}
\end{aligned}\end{equation} With the one-hot encoding of true class $n$, we define the class-wise CE loss w.r.t. source domain samples as \begin{equation}
\begin{aligned}
 \mathcal{L}_{CE}^{class}=-\text{log} p(y_i^s=n|x_i^s).
\end{aligned}\end{equation}The minimization of $\mathcal{L}_{CE}^{class}$ encourages the separation of classes in the source domain. {Therefore, the visually similar van and small truck in Fig. \ref{fig:sp1} are enforced to be different from each other, according to their class, which can be seen as a challenging case to facilitate the fine-grained feature exploration \cite{liu2019hard}.} The more explicit inter-class separation objective as in CAN \cite{kang2019contrastive} can potentially achieve further separation, leading to better performance.

Following the self-labeling scheme \cite{zou2019confidence,pan2019transferrable}, each target sample $x_i^t$ is assigned with a pseudo class label $\hat{y}_i^t$ to its nearest source centroids, i.e., \begin{equation}
\begin{aligned}
\hat{y}_i^t=n,~~\text{if}~~_{\forall n}^{\text{min}}||f(x_i^t)-c_n^s||_2^2.
\end{aligned}\end{equation} With the pseudo class label $\hat{y}_i^t$, we calculate the target domain class-level centroids \begin{equation}
\begin{aligned}
c_n^t=\frac{1}{M_c^t}\sum_{i=1}^{M_c^t}f(x_i^t),
\end{aligned}\end{equation} where ${M_c^t}$ is the number of target samples involved. We expect the close proximity of $c_n^s$ and $c_n^t$ with \begin{equation}
\begin{aligned}\label{rev}
\mathcal{L}^{class}={\frac{1}{N}\sum_{n=1}^N}||c_n^s-c_n^t||_2^2,
\end{aligned}\end{equation} which is not sensitive to label change, because it only selects the representative source and target distribution centroids\footnote{The source \& target center $c_n^{st}=\frac{\sum_{i=1}^{M_c^s+M_c^t}f(x_i^{st})}{M_c^s+M_c^t}$ used in \cite{pan2019transferrable}, and its objective of close proximity of $c_n^s\leftrightarrow c_n^{st}$, or $c_n^t\leftrightarrow c_n^{st}$ are not robust to label shift. Note that if we double the source/target samples involved, $c_n^{st}$ would shift.}. Neither $\mathcal{L}_{CE}^{class}$ nor $\mathcal{L}^{class}$, however, considers  the  fine-grained subtype structure and inner-class compactness \cite{wen2016discriminative} .

{We note that in class-wise separation and matching, the target class can be a subset of the source domains. We use the label to define the class center $c_n^s$ in the source domain, and assign the label to the target domain data with the nearest center. If there is no new class in the target domain, then the target sample should correspond to a proper class in the source domain. It is fine to have target samples that are labeled as some of the source classes (i.e., in the case where the target classes are a subset of source classes). In the case of no target domain sample being assigned to class $n$, we simply do not calculate the class matching loss $\mathcal{L}^{class}$ for class $n$ (see Eq. (\ref{rev})), similar to TPN.}

\subsection{Subtype-aware alignment with $K_n$ prior}



With the prior knowledge of subtype numbers $K_n$ of class $n$, we can achieve feature space class-independent clustering with the concise $K$-means \cite{coates2012learning}, by defining $K$ to be $K_n$.


We index $K_n$ clustered subtypes with $k\in\{1,2,\cdots, K_n\}$, and calculate the source subtype centroids $\mu_k^s$. After the $K_n$ subtypes in the source domain are explored and the corresponding centroids $\mu_k^s$ are calculated, we assign each target sample with pseudo-label of class $n$ to the subtype with the most adjacent centroid, i.e., \begin{equation}\begin{aligned}
x_i^t\in\text{subtype}~k~~\text{if}~~_{\forall k}^{\rm min}||f(x_i^t)-\mu_k^s||_2^2,\label{eq:7}
\end{aligned}\end{equation} which is used to calculate the target subtype centroids $\mu_k^t$.

Considering the imbalance distribution of subtypes and the possible label shift, we assign the subtype centroids of both source and target samples with \begin{equation}\begin{aligned}\label{rev2}
\mu_k^{st}=\frac{\mu_k^s+\mu_k^t}{2},
\end{aligned}\end{equation} instead of averaging all of the source and target samples in subtype $k$. By doing so, each subtype in both domains contributes equally to $\mu_k^{st}$. Then, we enforce all of the samples in subtype $k$ to be close to the subtype centroid $\mu_k^{s,t}$. We omit the class notation for simplicity, and the subtype compactness objective $\mathcal{L}_k^{sub}$ is given by \begin{equation}
\begin{aligned}
   \mathcal{L}_k^{sub}=\frac{1}{M_k^s}\sum_{i=1}^{M_k^s}||f(x_i^s)-\mu_k^{st}||_2^2+\frac{1}{M_k^t}\sum_{i=1}^{M_k^t}||f(x_i^t)-\mu_k^{st}||_2^2, \label{eq:2}
\end{aligned}\end{equation} where $M_k^s$ and $M_k^t$ are the source and target sample numbers of subtype $k$ to balance the subtype label shift. $\mathcal{L}_k^{sub}$ is traversed for $N$ classes and their $K_n$ subtypes to calculate the sum of normalized subtype compactness losses \begin{equation}\begin{aligned}
\mathcal{L}^{sub}=\frac{1}{N}\sum^N(\frac{1}{K_n}\sum^{K_n}\mathcal{L}_k^{sub}).
\end{aligned}\end{equation} Of note, an image that does not belong to class $n$ does not belong either to any of its $K_n$ subtypes. The class-wise matching and subtype-wise compactness objectives can be aggregated as a hierarchical alignment loss as \begin{equation}\begin{aligned}
\frac{1}{N}\sum^N(\alpha\mathcal{L}^{class}+\beta\frac{1}{K_n}\sum^{K_n}\mathcal{L}_k^{sub}),
\end{aligned}\end{equation} where $\alpha$ and $\beta$ are balancing parameters. In the feature space, the learned representations are expected to form $K_n$ compact clusters for class $n$, although each cluster does not need to be far apart from one another. 

{We note that in our sub-type clustering, the target subtypes can also be a subset of the source domains. In both $K$-means and reliability-path-based sub-graph, we cluster the source domain data first. Then, we assign the subtype of the target domain samples to the nearest source domain subtype center (see Eq.(\ref{eq:7})) just like the pseudo-class label assignment as discussed above. Therefore, there can be no target data for a subtype in the source domain, and we have $\mu_k^st=\mu_k^s$. Consequently, Eq. (\ref{rev2}) enforces the compactness of the source domain data in this subtype only. In our reliability-path-based sub-graph construction, we do not even define the number of subtypes.}


\subsection{Reliability-path based sub-graph construction}


{In practice, extensive trials are required to accurately estimate $K_n$ for $n$ classes, because of the varying value ranges of different classes, which is impractical especially for large scale tasks.} Notably, however, with a deterministic encoder $f$, the distribution protocol can be similar among different subtypes and classes \cite{fahad2014survey}.

In order to achieve the online source domain subtype clustering, we propose to form a sub-graph with a reliability-path. We presume that with a deterministic encoder $f$, similar samples are likely to be closely distributed in the latent space and form a high-density region \cite{carlucci2019domain}. Given $M^{s}$ samples from class $n$ in the source domain, there are $(M^{s})^2$ possible edges in a graph. To mine the local structure inherent in the feature space,  two nodes $\{f(x_i^s),f(x_j^s)\}$ are connected by the reliability-path if $||f(x_i^s)-f(x_j^s)||_2^2\leq\epsilon$. To form a sub-graph, the directly or indirectly linked nodes are combined.

To further eliminate the effect of noise and undersampled subtypes on a batch, we only select the sub-graphs with more than $m$ nodes as valid subtype clusters. After $K_n$ subtypes in the source domain are explored, and the corresponding centroids $\mu_k^s$ are calculated, we assign each target sample with pseudo-label of class $n$ to the subtype with the most adjacent centroid as in Eq. (\ref{eq:7}).

\begin{figure}[t]
\centering
\includegraphics[width=8.5cm]{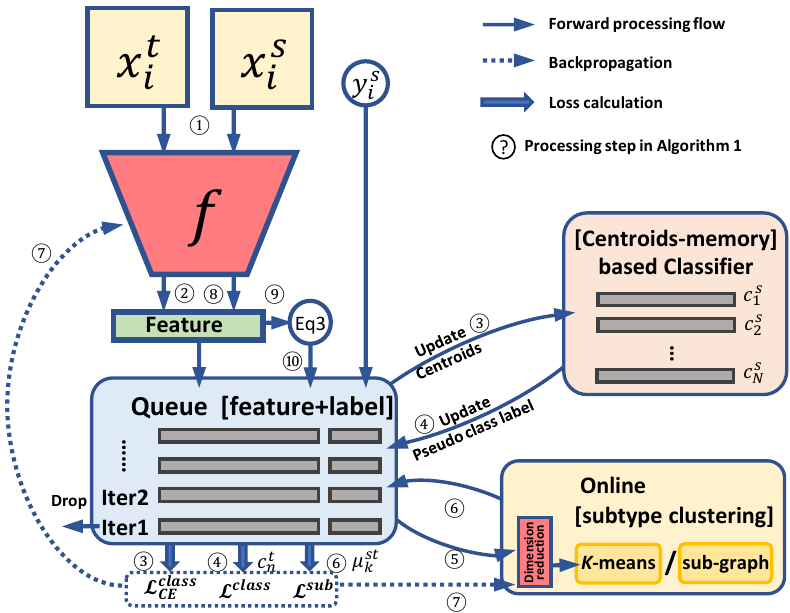}\\ 
\caption{Illustration of our dynamic online clustering framework with an alternatively updated feature queue module and a centroids memory module.}\label{fig:44} 
\end{figure}

Since the confidence or reliability of pseudo target labels is relatively low  \cite{zou2019confidence,gu2020spherical}, we propose to apply a simple online semi-hard mining scheme to choose the target sample in a subtype. The cross-domain margin $\tau$ is used to define a circle at the center of $\mu_k^s$. 

Given the target samples with the initial pseudo subtype label $k$, we only select the samples that are distributed within the circle. Considering that some target samples may be densely distributed around the circle boundary, it is not reasonable to simply cut them apart. Therefore, we also resort to the reliability-path to incorporate the closely distributed neighboring target samples. The sub-graph construction can be robust to missing subtypes in the source or target domain caused by undersampling, because $m$ filters the unreliable source cluster out, and the self-labeling with semi-hard mining of $f(x_i^t)$ rejects the additional subtypes in the sampled target domain. We illustrate the operations in Fig. \ref{fig:33}.

With the reliability-path connected $M_k^s$ source samples and the refined $M_k^t$ target samples in subtype $k$, we calculate $\mu_k^{st}=\frac{\mu_k^s+\mu_k^t}{2}$, and enforce the subtype-wise compactness with $\mathcal{L}_k^{sub}$ as in Eq. (\ref{eq:2}). 

In summary, there are three hyperparameters in our online sub-graph construction, including $\epsilon$, $\tau$, and $m$. They are shared for all classes and their subtypes, which can be regarded as meta-knowledge across clusters. Furthermore, $\epsilon$ can be simplified to constant 1. We can even change it to any positive number via multiplying the corresponding factors \cite{liu2017adaptive}. The range of $m$ can also be narrow and similar among different subtypes/classes.

\subsection{Queued dynamic clustering}

There is a possibility that there could be many underlying subtypes, i.e., $K_n$ is large, and some of the subtypes may be undersampled in both domains of a mini-batch. As such, it is challenging to obtain the representative subtype-wise sampling with a small batch. Therefore, instead of extracting the feature of the entire training set at each iteration via deep clustering \cite{caron2018deep}, which requires extensive computation for unsupervised clustering and large GPU memory (if the dataset is very large, e.g., DomainNet), we propose to incorporate more representative samples with a dynamic feature queue module that evolves smoothly along with the network parameter update.

\begin{algorithm}[t]
\caption{Dynamic SubUDA}
\label{alg:A}
\begin{algorithmic}
\STATE {Initializing network parameters} 
\REPEAT 
\STATE 1. Sample the training batch $\left\{x_i^s,y_i^s\right\}$, $\left\{x_i^t\right\}$ 
 
\STATE 2. Forward pass to extract $\left\{f(x_i^s)\right\}$ and $\left\{f(x_i^t)\right\}$  

\STATE 3. {$c_n^s$ of $\mathcal{I}=5$ batches} and calculate $\mathcal{L}_{CE}^{class}$ of $\left\{f(x_i^s)\right\}$
\STATE 4. Self-label $\left\{f(x_i^t)\right\}$ and calculate $\mathcal{L}^{class}$ with $c_n^t$
\STATE //Subtype clustering and matching to compute $\mu_k^{s}$, $\mu_k^{t}$
\STATE 5. Perform $K$-means or reliability-path based sub-graph 

\STATE 6. $\mu_k^{st}\leftarrow\frac{1}{2}(\mu_k^{s}+\mu_k^{t})$, and calculate $\mathcal{L}_{k}^{sub}$

\STATE//Update parameters according to gradients
\STATE 7. $f\leftarrow \mathcal{L}_{CE}^{class}+\frac{1}{N}\sum^N(\alpha\mathcal{L}^{class}+\beta\frac{\omega_k}{K_n}\sum^{K_n}\mathcal{L}_k^{sub})$
\STATE 8. Re-compute $\left\{f'(x_i^s)\right\}$ and $\left\{f'(x_i^t)\right\}$ 
\STATE 9. Store $f''(x_i^{st})\leftarrow \lambda f'(x_i^{st}) + (1-\lambda)f(x_i^{st})$ to queue

\STATE 10. Update centroids memory module.

\UNTIL{deadline}
\end{algorithmic}
\end{algorithm}

\begin{figure*}[t]
\begin{center}
\includegraphics[width=1\linewidth]{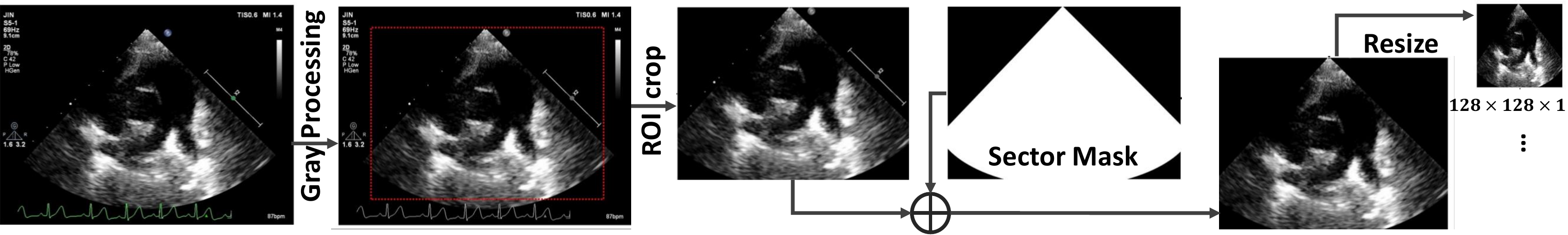}
\end{center} 
\caption{Prepossessing of the CHD task.}
\label{fig:2}
\end{figure*}

As shown in Fig. \ref{fig:44}, {the dynamic queue maintains the features and their label of $\mathcal{I}$ iterations, where $\mathcal{I}$ is usually set to 5 in our implementation. With more sampled data in multiple batches, the calculated center can be more representative.} Moreover, the centroids memory module stores the class and subtype centers with the current $\mathcal{I}$ batches. The ``subtype'' here represents the temporary cluster that evolves continuously along with the training iterations. Our centroid-based classifiers and networks are updated simultaneously during uninterrupted iterations. The detailed processing is given in Algorithm \ref{alg:A}.

{In our dynamic memory scheme, we store 5 batches samples, and update the batches along with the training. The batch size is set to 64 in all of our tasks. With Float32 in Pytorch, each dimension takes 4 Bytes. Therefore, the memory cost of the 256-dim feature of 5 batches samples is 64$\times$5$\times$256$\times$4 = 327,680 Bytes = 327.68 KB. Compared with the memory of our GPU device, i.e., 32GB, the memory cost of the feature storage is relatively low. The I/O can also be compared with the unsupervised clustering. In fact, the unsupervised clustering, e.g., $K$-means, limits the scale of batch in our SubUDA. Our dynamic memory does follow the first-in-first-out (FIFO) protocol to keep 5 recent batches along with the iterations. As shown in Fig. \ref{fig:44}, the early batches are dropped. Thus, the queue is a more natural data structure choice.}\footnote{{We used ``collections.deque" to construct the queue in Python, which can efficiently add or remove the elements on both sides of the queue.}}

With a batch of training samples $\{x_i^s\}$, $\{x_i^t\}$ in an iteration, we first encode these samples into feature vectors $\{f(x_i^s)\}$, $\{f(x_i^t)\}$, which are added to the queue. Then, we read the label of source samples $\{x_i^s\}$ in this batch to calculate the class-level CE loss $\mathcal{L}_{CE}^{class}$. The pseudo-label of target samples is self-labeled with the centroids of source features in the $\mathcal{I}$ batches, and $\mathcal{L}^{class}$ is also applied to the source and target class centroids among $\mathcal{I}$ batches.

After the class-wise pseudo-label is inferred, we can explore the subtype structure of paired class-wise source and target distributions on $\mathcal{I}$ batches. Based on the availability of the subtype numbers, $K$-means or online reliability-path based sub-graph construction is applied to adaptively cluster $\mathcal{I}$ batches; therefore, $\mathcal{L}^{sub}$ can be calculated. The overall optimization objective can be expressed as\begin{equation}
\begin{aligned}
\mathcal{L}=\mathcal{L}_{CE}^{class}+\alpha\mathcal{L}^{class}+\beta\mathcal{L}^{sub},
\end{aligned}
\end{equation} and we update $f$ with stochastic gradient descent.

With the updated parameters, we re-compute the feature representation of this batch $f'(x_i^{st})$. The feature memory in queue is dynamically rewritten with a momentum strategy:\begin{equation}
\begin{aligned}
f''(x_i^{st})\leftarrow \lambda f'(x_i^{st}) + (1-\lambda)f(x_i^{st}),
\end{aligned}
\end{equation} where $\lambda$ is an empirical momentum weight, and $f(x_i^{st})$ and $f'(x_i^{st})$ are the feature representations extracted by the original and updated extractor in this iteration, respectively. Note that $f''(x_i^{st})$ of this batch is stored in our dynamic queue.

\subsection{Optimization and implementation}

\begin{figure}[t]
\begin{center}
\includegraphics[width=1\linewidth]{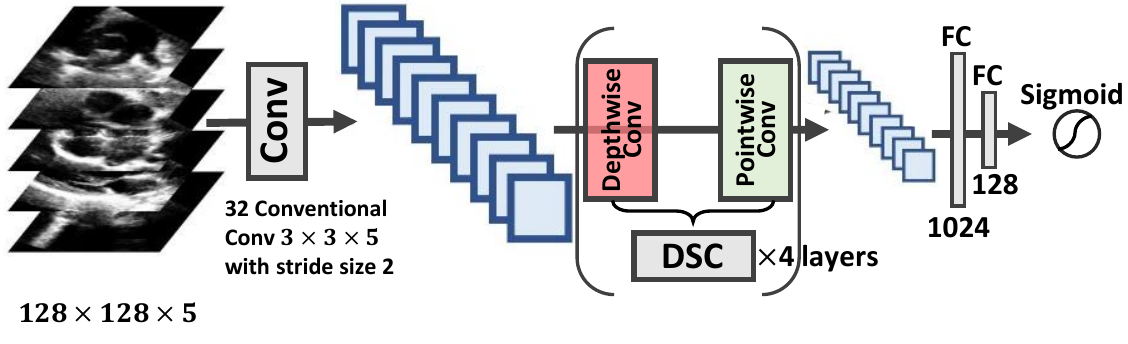}
\end{center} \vspace{-10pt}
\caption{The multichannel DSC backbone for our five-view CHD analysis.}
\label{fig:4}
\end{figure}

Many backbone networks extract high-dimensional representations, such as 4,096 or 2,048-dimensional features, from image datasets, thereby demanding high time cost in subsequent clustering and memory cost w.r.t. memory. To alleviate this, deep clustering \cite{caron2018deep} performs Principal Component Analysis (PCA) on whole datasets for dimension reduction. However, in our online SubUDA, PCA is no longer applicable, as there are different timestamps of differently-sampled features, leading to incompatible sample statistics. 

Applying PCA in each iteration is also computationally expensive. Therefore, to compact high-dimensional features to 256 dimensions, we simply add a non-linear head-layer of {fc$\rightarrow$bn$\rightarrow$relu$\rightarrow$dropout$\rightarrow$fc$\rightarrow$relu} jointly tuned during online SubUDA iterations. For downstream tasks, such as calculating the L2 distance between features, the head layer is removed.

\begin{table}[t]
\caption{{The detailed structure of our five-channel DSC network. The Conv dw/pw indicate the depth and point wise convolutions, respectively. The 1st-layer adopts the traditional convolutional kernel (stride is set to 2) \cite{howard2017mobilenets}.}} 
\centering 
\resizebox{1\columnwidth}{!}{%
\begin{tabular}{l | l | l} 
\hline\hline 
Input Size&Type / Stride & Filter Shape   \\ [0.5ex] 
\hline 

$128\times128\times5$&Conv / s2 & 32 kernels of $3\times3\times5$  \\
\hline

$64\times64\times32$&Conv dw / s1& 32 kernels of $3\times3$ dw \\

$64\times64\times32$& Conv pw / s1& 64 kernels of$1\times1\times32$ pw \\
\hline

$64\times64\times64$& Conv dw / s2& 64 kernels of$3\times3$ dw \\

$32\times32\times64$& Conv pw / s1& 128 kernels of$1\times1\times64$ pw \\
\hline

$32\times32\times128$& Conv dw / s2& 128 kernels of $3\times3$ dw  \\

$16\times16\times128$&Conv pw / s1&128 kernels of $1\times1\times128$ pw  \\
\hline

$16\times16\times128$&Conv dw / s2&128 kernels of $3\times3$ dw \\

$8\times8\times128$& Conv pw / s1&128 kernels of $1\times1\times128$ pw \\
\hline

$8\times8\times128$ & Flatten & N/A\\

8192& FC1 & 1024 \\

1024& FC2 & 128 \\
\hline

128& Classifier & Sigmoid \\
\hline
 
\end{tabular}
\label{Table:mobilenet} 
}
\end{table}

\cite{caron2018deep} conducts uniform sampling in all of the training epochs to prevent the training from collapsing into a few large subtype clusters. However, without the subtype and target class labels, it is difficult to apply the same strategy in our online UDA environment. Thus, we develop a concise method for SubUDA to re-weight the loss function with\begin{equation}
\begin{aligned}
\omega_k\propto\frac{1}{\sqrt{M_k^s+M_k^t}},
\end{aligned}
\end{equation} based on the sample numbers in the $k$-{th} subtype. Samples from smaller clusters tend to contribute more to backpropagation. Therefore, in order to include more potential samples, the decision boundary tend to move farther. We can summarize the optimization objective as \begin{equation}
\begin{aligned}
\mathcal{L}=\mathcal{L}_{CE}^{class}+\frac{1}{N}\sum^N(\alpha\mathcal{L}^{class}+\beta\frac{\omega_k}{K_n}\sum^{K_n}\mathcal{L}_k^{sub}).
\end{aligned}
\end{equation}

As stated above, the subtype degenerates to class alignment, if we set the subtype number of each class as 1. If we know the subtype number $K_n$ and use $K$-means, there would be no such kind of problem. In some evaluations, we just set $K_n=1$ to illustrate the necessity for exploring subtypes. For the case of unknown $K_n$, using relatively small $\tau$ and $m$ can also be helpful to yield many sub-clusters.

We note that the online clustering is performed in every batch iteration, and we only calculate with the samples only in a batch and compare their feature with the previous center vector, rather than all of the previous features. For the classification in testing, the centroids of training features are used as prototypical \cite{pan2019transferrable}.  
 
\begin{figure}[t]
\centering
\includegraphics[width=9cm]{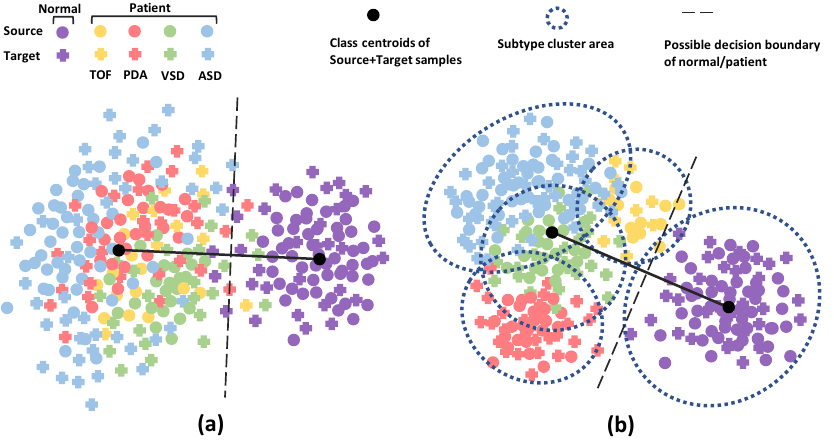}\\ 
\caption{Visualization of the sampled CHD features with t-SNE, using (a) TPN \cite{pan2019transferrable} and (b) our SubUDA. We normalized the distance between class centroids to demonstrate the class separation and inner-class/subtype compactness.}\label{fig:55} 
\end{figure}

\section{Experiments}

Our experiments were performed both on medical subtype data and the popular VisDA and DomainNet benchmarks to evaluate the effectiveness of our framework. We implemented our framework using PyTorch and the reported training times were based on a V100 GPU (32GB memory). 

{For the class-level prototypes used in testing, we did not see a significant difference. After the training, the network was able to embed these three prototypes closely as TPN \cite{pan2019transferrable}. As shown in Fig. \ref{tabel:abla}, the difference is consistently smaller than 0.002. The similar performance also implicitly demonstrates the domain invariant property of our extracted feature representations. The reported results in the subsequent evaluations are based on the prototypes of $c_n^{s,t}$.}

{For a fair comparison, we follow the previous standard evaluation protocol to validate the hyper-parameters with grid searching. The sensitivity analysis of these hyper-parameters (e.g., $K_n$, m, $\tau$, $\epsilon$) is also provided in Figs. 9-13.}

\begin{table}[t]
\centering
\resizebox{1\linewidth}{!}{%
\begin{tabular}{l|cccccc|c}
\hline

Method for CHD & prototypes in test & Accuracy (\%) $\uparrow$    \\ \hline

SubUDA ($K_n=4$)&$c_n^{s,t}$ & {96.8$\pm$0.13}  \\\hline
 
SubUDA ($K_n=4$)&$c_n^{s}$ &  {96.8$\pm$0.14}  \\ 
 
SubUDA ($K_n=4$)&$c_n^{t}$ &  {96.7$\pm$0.12}  \\\hline\hline

Method for VisDA17& prototypes in test & Mean Acc. (\%) $\uparrow$  \\ \hline

SubUDA ($m=5$)&$c_n^{s,t}$ & 82.8$\pm$0.4    \\

SubUDA ($m=5$)&$c_n^{s}$ & 82.7$\pm$0.3    \\

SubUDA ($m=5$)&$c_n^{t}$ & 82.4$\pm$0.4     \\ \hline\hline

\end{tabular}%
}
 
\caption{{Ablation studies using different prototypes in SubUDA for CHD and VisDA17 testing.}}
 
\label{tabel:abla}
\end{table} 

\begin{table}[t]
\centering
\resizebox{1\linewidth}{!}{%
\begin{tabular}{l|cccccc|c}
\hline

Compared Methods & Accuracy (\%) $\uparrow$ & AUC $\uparrow$  \\ \hline

Source only& 76.4$\pm$0.12  & 0.721$\pm$0.005 \\

MCD \cite{saito2017maximum}& 88.6$\pm$0.15  & 0.856$\pm$0.003 \\

GTA \cite{sankaranarayanan2018generate}& 90.9$\pm$0.17  & 0.873$\pm$0.005 \\

CRST \cite{zou2019confidence}& 93.2$\pm$0.09  & 0.882$\pm$0.006 \\

Prototypical Network (TPN) \cite{pan2019transferrable}& 93.4$\pm$0.14  & 0.885$\pm$0.004 \\\hline\hline

Proposed Methods & Accuracy (\%) $\uparrow$ & AUC $\uparrow$  \\ \hline

{\textbf{SubUDA}+Dyn} ($K_n=4$)& 96.8$\pm$0.16  & 0.921$\pm$0.002 \\

{\textbf{SubUDA} ($K_n=4$)\cite{Liu2021subtype}}&  {96.2$\pm$0.13}  &  {0.910$\pm$0.003} \\ 

\textbf{SubUDA} ($K_n=1$):class-wise compactness&  {94.7$\pm$0.11}  &  {0.902$\pm$0.004} \\

\textbf{SubUDA}-$\mu^{st}_k$ ($K_n=4$)& 95.4$\pm$0.10  &  0.903$\pm$0.005\\
\textbf{SubUDA}-$\omega_k$ ($K_n=4$)& 96.0$\pm$0.13  &  0.908$\pm$0.004\\
\textbf{SubUDA}-DR ($K_n=4$)&  96.2$\pm$0.11 &  0.911$\pm$0.002\\
\hline

\textbf{SubUDA}+Dyn ($m=8$)& 96.6$\pm$0.14  & 0.914$\pm$0.005\\

{\textbf{SubUDA} ($m=8$)\cite{Liu2021subtype} }& 96.0$\pm$0.12  & 0.907$\pm$0.004  \\

\textbf{SubUDA}-$\tau$ ($m=8$)& 95.5$\pm$0.14  & 0.902$\pm$0.003 \\\hline

\end{tabular}%
}
 
\caption{Experimental results for CHD. $\uparrow$ larger is better. Our proposed SubUDA and its variants are bold.}
 
\label{tabel:chd}
\end{table}

\subsection{CHD Transfer Task}

Congenital heart defects (CHDs) are the most common congenital disability, which leads to the neonates' death. Usually, the clinical diagnosis of the disease relies on chosen echocardiograms from multi-views. A multi-center dataset of five-view echocardiograms of 1,608 labeled source subjects (normal/patient) from Beijing Children's Hospital (BCH) and 800 unlabeled target subjects from the Harvard Medical School (HMS) was acquired. Note that there were discrepancies w.r.t. imaging devices (PHILIPS iE 33 vs. EPIQ 7C), patient populations, and the echocardiogram imaging experience of clinicians, which collectively introduced the domain shift. The research protocol was granted by the Ethics Committee of BCH (No. 2019-k-342).

The four subtypes of CHD, including atrial septal defect (ASD), ventricular septal defect (VSD), patent ductus artery (PDA), and tetralogy of Fallot (TOF), were confirmed by at least two senior ultrasound clinicians or by a final intraoperative diagnosis to analyze the influence of the subtype structure. We note that the fine-grained subtype label was not used in training, as large-scale labeling was expensive, whereas primary clinicians can obtain the normal/patient label relatively easily. In addition, we attempted to tackle the subtype-aware alignment for the conventional class-wise discriminative model. Fig. \ref{fig:55} left shows that a subtype's source CHD samples often appear to be tightly distributed, indicating the underlying similarity of the inner-subtype. With our subtype-wise compactness objective, both the source and target samples were grouped into high-density regions w.r.t. subtypes. 

\subsubsection{Data collection and prepossessing}

We placed each subject in the supine position and exposed the chest area to the echocardiogram. In total, five standard 2D views were acquired, including parasternal long-axis (PSLAX), parasternal short-axis (PSSAX), apical four chambers (A4C), subxiphoid long-axis (SXLAX), and suprasternal long-axis (SSLAX). The key-frame was manually selected by a skilled clinician. In particular, we marked the isovolumic relaxation process as a key-frame, if the contraction of ventricles is stopped.

\begin{figure}[t]
\centering
\includegraphics[width=9cm]{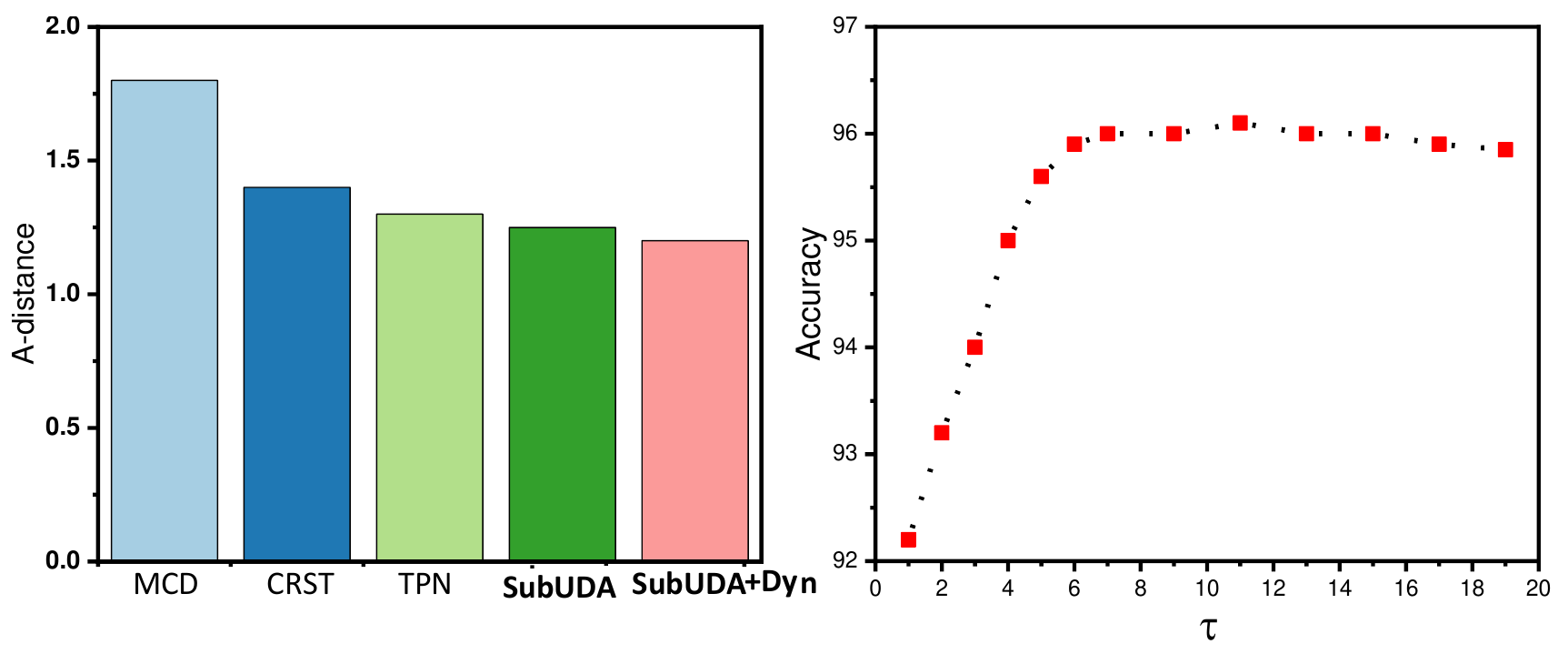}\\ 
\caption{Comparison with the other STOAs w.r.t. $\mathcal{A}$-distance (left), and the sensitive analysis of $\tau$ (right) in SubUDA for CHD task. We use the discriminator as \cite{saito2017adversarial}.}\label{fig:66} 
\end{figure}

A total of 1,608 subjects in the BCH source domain (823 normal, 209 VSD, 276 ASD, 124 TOF, and 176 PDA) were used in this work. For the acquisition, we used PHILIPS iE 33. The frequency of the transducer ranged between 3 and 8 MHz. A total of 800 subjects were collected in the HMS target domain (300 normal, 150 VSD, 150 ASD, 100 TOF, and 100 PDA). For the acquisition, we used PHILIPS EPIQ 7C. The frequency of the transducer ranged between 3 and 8 MHz. 

In sum, imaging instruments (PHILIPS iE 33 vs. EPIQ 7C), patient demographics, and the echocardiogram imaging background of healthcare clinicians were the major differences between these two medical centers, which caused the data shifts between two domains. We used 75\%/5\%/20\%  split of the target subjects for training, validation and testing.

\subsubsection{Backbone network structure for the CHD task}

The effectiveness of Depthwise Separable Convolution (DSC) \cite{howard2017mobilenets} based multi-channel network for single domain five-view echocardiogram processing has been successfully demonstrated \cite{Liu2021Automated}. Instead of merely concatenating in a fully connected layer as in multi-branch networks, the multi-channel scheme was able to adaptively fuse the knowledge in all layers \cite{lee2016multi}. We stacked the five views sequentially to form a matrix with the size of 128$\times$128$\times$5. As shown in Fig. \ref{fig:4}, we adopted a five-channel convolutional neural network to process the samples. The DSC backbone \cite{howard2017mobilenets} can significantly reduce the number of to-be-learned parameters, which can be helpful for small datasets \cite{lee2016multi}. While AlexNet used 60 million parameters in training \cite{simonyan2014very}, the DSC (with width multiplier 0.50) \cite{howard2017mobilenets} achieved a similar accuracy on ImageNet with much fewer parameters (about 1.32 million parameters). A block of DSC combined a depth and point wise convolution. The feature maps were flattened after a few DSC layers, followed by two fully connected (FC) layers with the sizes of 1024 and 128, respectively. We provide the network details in Table \ref{Table:mobilenet}. We used the sigmoid as the output layer for binary classification (normal or patient) and optimized the network with the binary CE loss. {We set $\alpha=1$, $\lambda=0.5$, and $\beta=1$ and used Adam with lr=1e-5.}

\subsubsection{Evaluations}

For a fair comparison, we re-implemented the state-of-the-art methods, following the same evaluation protocol and using the same backbone. We empirically set the batch size to 64 and $\mathcal{I}=5$. Table \ref{tabel:chd} shows our results, where we reported the accuracy with the threshold of 0.5. Given the normal and patient proportion disparity in the testing set, in addition to accuracy, the area under the receiver operating characteristic curve (AUC) is included, which is a threshold-irrelevant metric.

\begin{figure}[t]
\centering
\includegraphics[width=9cm]{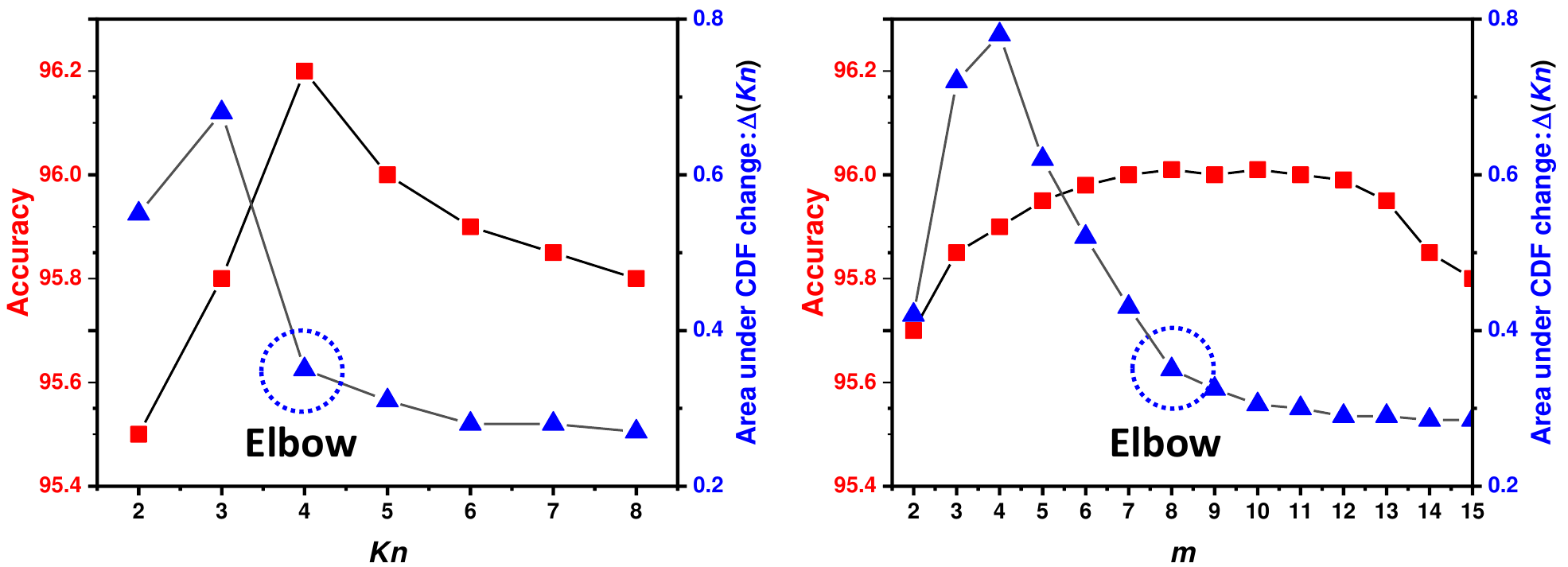}\\ 
\caption{Sensitive analysis of $K_n$ (right) and $m$ of patient class in CHD dataset with SubUDA. The red rectangle indicates accuracy and the blue triangle indicates AUC of CDF. }\label{fig:77} 
\end{figure}

Among these works, MCD \cite{saito2017maximum} and GTA \cite{sankaranarayanan2018generate} were the typical adversarial training frameworks. They focused on feature or image-level alignment of the marginal distribution $p(x)$. Besides, the self-training based on \cite{zou2019confidence} utilized the alternative training to simultaneously update the pseudo-label of target samples and network parameters. We note that the fully-connected classifier after the encoder was adopted in \cite{saito2017maximum,sankaranarayanan2018generate,zou2019confidence}, while TPN \cite{pan2019transferrable} utilized the class centroids as its classifiers. The TPN aimed to achieve the conditional alignment w.r.t. $p(x|y)$ via matching the class-wise centroids among two domains. In contrast, our SubUDA outperformed the comparison methods w.r.t. both the accuracy and AUC, by introducing the subtype-aware constraint. The performance clearly indicates the effectiveness of the online subtype compactness for the target domain classification.

According to the domain adaptation theory \cite{ben2007analysis}, the proxy $\mathcal{A}$-distance \cite{saito2017asymmetric} can be used to measure the cross-domain discrepancy. Accordingly, our proposed SubUDA was compared with the other methods using the distance as shown in Fig. \ref{fig:66}. The SubUDA achieved a smaller discrepancy w.r.t. feature representation of source and target domains via an explicit compactness objective.

For the ablation study w.r.t. $K_n=1$, the subtype-wise alignment degenerated to the class-wise compactness. In addition, the improvement of SubUDA \cite{Liu2021subtype} with $K_n=1$ over TPN is contributed by enforcing the class-wise compactness of both source and target samples and the balanced label shift. We used suffixes +Dyn to denote dynamic subUDA with dynamic queue memory. Also, -DR, -$\omega_k$, and -$\tau$ indicate subUDA without dimension reduction head, subtype balance weight, and semi-hard target mining, respectively. {We note that the SubUDA was proposed in our prior work \cite{Liu2021subtype}, which did not include the dynamic memory scheme.}

In addition, the suffix -$\mu_k^{st}$ denots using $\mu_k^{st}=\frac{\sum_{i=1}^{M_k^s+M_k^t}f(x_i^{st})}{M_k^s+M_k^t}$ as the subtype centroid, which is not robust against the subtype label shift. Among them, the dynamic queue scheme played the most important role in addressing the difficulty of subtype undersampling and subtype label shifts. SubUDA-DR took 4$\times$ more clustering time, while the improvement was marginal. Therefore, we would recommend using a dimension reduction head to achieve more efficient processing. 

\begin{figure}[t]
\centering
\includegraphics[width=9cm]{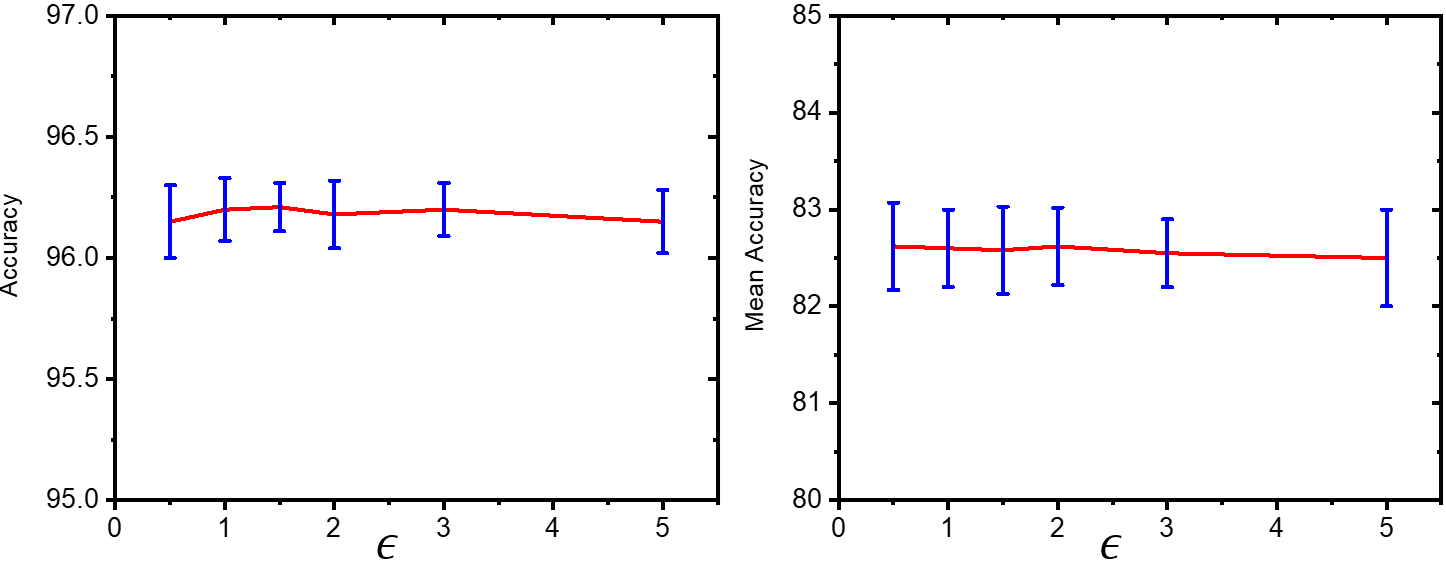}\\ 
\caption{{Sensitive analysis of $\epsilon$ in SubUDA for CHD and VisDA17 datasets.}}\label{fig:abla} 
\end{figure}

\begin{table*}[!t]
\centering
\resizebox{\linewidth}{!}{
\centering
\begin{tabular}{l|cccccccccccc|c}
\hline
Method         &  Aero & Bike   & Bus & Car & Horse & Knife & Motor   & Person   & Plant & Skateboard & Train  & Truck   & Mean \\ \hline\hline
ResNet101(source only) & 55.1 & 53.3 & 61.9 & 59.1 & 80.6 & 17.9 & 79.7 & 31.2 & 81.0 & 26.5 & 73.5 & 8.5 & 52.4 \\\hline

MMD (ICML15) \cite{long2015learning} & 87.1 & 63.0 & 76.5 & 42.0 & 90.3 & 42.9 & 85.9 & 53.1 & 49.7 & 36.3 & 85.8 & 20.7 & 61.1 \\

DANN (JMLR16) \cite{ganin2016domain} & 81.9 & 77.7 & 82.8 & 44.3 & 81.2 & 29.5 & 65.1 & 28.6 & 51.9 & 54.6 & 82.8 & 7.8 & 57.4 \\

MCD (CVPR18) \cite{saito2017maximum}  & 87.0 & 60.9 & {83.7} & 64.0 & 88.9 & 79.6 & 84.7 &  {76.9} &  {88.6} & 40.3 & 83.0 & 25.8 & 71.9 \\

ADR (ICLR18) \cite{saito2017adversarial}  & 87.8 & 79.5 & {83.7} & 65.3 & {92.3} & 61.8 &  {88.9} & 73.2 & 87.8 & 60.0 &  {85.5} & {32.3} & 74.8 \\  

DEV (ICML19) \cite{you2019toward}  & 81.8& 53.5& 83.0& 71.6& 89.2 &72.0& 89.4& 75.7& {97.0} & 55.5 &71.2 &29.2& 72.4\\

CRST (ICCV19) \cite{zou2019confidence}  & 89.2 &79.6 &64.2 &57.8 &87.8 &79.6 &85.6 &75.9 &86.5 &85.1 &77.7 &68.5 &78.1\\
 
PANDA (ECCV20) \cite{hu2020panda} &90.9 &50.5 &72.3& 82.7 &88.3 &88.3 &90.3& 79.8& 89.7 &79.2& 88.1 &39.4& 78.3\\

DMRL (ECCV20) \cite{wu2020dual} & - &-& -&- &- &-& - &- &-& -& -& - &75.5\\

MEDM (TNNLS21) \cite{wu2021entropy} &93.5 &80.4 &90.8 &70.3& 92.8 &87.9 &91.1& 79.8& 93.7& 83.6 &86.1 &38.7 &82.4\\

CLS (ICCV21) \cite{liu2021adversarial} & 92.6$\pm$1.5 & 84.5$\pm$1.4 & 73.7$\pm$1.5 & 72.7$\pm$1.1 & 88.5$\pm$1.4 & 83.3$\pm$1.2 & 89.1$\pm$1.4 & 77.6$\pm$1.5 & 89.5$\pm$1.5 & 89.2$\pm$1.3 & 85.8$\pm$1.2 & 72.7$\pm$1.3 & 81.6$\pm$0.4\\

TPN (CVPR19) \cite{pan2019transferrable}  &  93.7 &  85.1 &  69.2  & 81.6  & 93.5  & 61.9  & 89.3 &  81.4  & 93.5 &  81.6  & 84.5  & 49.9  & 80.4\\

\hline

\textbf{SubUDA}+Dyn ($m=5$) & {93.8$\pm$0.8} & {85.5$\pm$0.7} & 75.1$\pm$0.5 & {73.9$\pm$0.4} & 91.3$\pm$0.6 & {83.8$\pm$0.7} & 89.7$\pm$0.6 & {82.3$\pm$0.5 }& 90.4$\pm$0.4 & {91.5$\pm$0.7} & {87.2$\pm$0.6} & {74.0$\pm$0.5 }& {82.8$\pm$0.4} \\\hline

{\textbf{SubUDA} \cite{Liu2021subtype}} ($m=5$)&  92.8$\pm$0.8 & 85.4$\pm$0.9 & 74.3$\pm$0.7 & 72.5$\pm$0.8 & 88.2$\pm$0.9 & 81.8$\pm$0.7 & 86.6$\pm$1.0 & 77.3$\pm$0.6 & 86.6$\pm$1.0 & 89.5$\pm$0.7 & 86.7$\pm$0.6 & 72.0$\pm$0.8 & 81.8$\pm$0.6 \\

\textbf{SubUDA}-$\omega_k$ ($m=5$)& 93.0$\pm$0.5 & 85.2$\pm$0.3 & 73.8$\pm$0.7 & 73.5$\pm$0.6 & 88.6$\pm$0.5 & {83.7$\pm$1.0} & 87.4$\pm$0.6 & 78.0$\pm$0.4 & 87.3$\pm$0.7 & 90.6$\pm$0.5 & 86.5$\pm$0.6 & 73.1$\pm$0.5 & 82.3$\pm$0.4\\

\textbf{SubUDA}-$\mu^{st}_k$ ($m=5$)& 92.9$\pm$0.8 & 84.6$\pm$0.5 & 73.5$\pm$0.6 & 73.6$\pm$0.8 & 88.2$\pm$0.6 & {83.4$\pm$0.7} & 88.5$\pm$0.5 & 77.2$\pm$0.5 & 87.2$\pm$0.8 & 89.3$\pm$0.5 & 85.9$\pm$0.7 & 72.5$\pm$0.6 & 82.0$\pm$0.5\\

\textbf{SubUDA}-$\tau$ ($m=5$)& 92.8$\pm$0.5 & 84.9$\pm$0.7 & 74.0$\pm$0.5 & 73.1$\pm$0.4 & 88.8$\pm$0.6 & 83.5$\pm$0.7 & 89.4$\pm$0.7 & 78.0$\pm$0.5 & 89.7$\pm$0.6 & 89.6$\pm$0.4 & 85.9$\pm$0.6 & 72.8$\pm$0.5 & 81.9$\pm$0.4\\

\textbf{SubUDA}-$K_n$ (validated $K_n$)& 93.6$\pm$0.7 & {85.5$\pm$0.8} & 74.7$\pm$0.6 & 73.4$\pm$0.5 & 91.3$\pm$0.6 & 83.2$\pm$0.7 & {91.2$\pm$0.5} & 81.7$\pm$0.6 & 89.1$\pm$0.7 & 89.9$\pm$0.5 & {87.2$\pm$0.7} & 73.1$\pm$0.8 & 82.7$\pm$0.6\\

\textbf{SubUDA}-$K_n$ ($K_n=1$)& 91.2$\pm$0.5 & 83.7$\pm$0.6 & 73.5$\pm$0.7 & 72.6$\pm$0.9 & 88.7$\pm$0.5& {81.4$\pm$0.6} & 87.3$\pm$0.4 & {81.9$\pm$0.5} & 88.2$\pm$0.8 & {90.8$\pm$0.7} & 84.5$\pm$0.6 & 72.6$\pm$0.5 & 81.2$\pm$0.5\\\hline\hline

{CAN}(CVPR19) \cite{kang2019contrastive} & 97.0 &87.2& 82.5& 74.3& 97.8& 96.2& 90.8& 80.7& 96.6& 96.3& 87.5& 59.9& 87.2 \\ 

{CAN + A2LP}(ECCV20)  \cite{zhang2020label} & 97.5& 86.9 &83.1 &74.2& 98.0 &97.4 &90.5 &80.9 &96.9 &96.5 &89.0& 60.1& 87.6 \\\hline

{CAN + \textbf{SubUDA}} ($m=5$) &  {97.8$\pm$0.4} &  {85.6$\pm$0.5} & 75.2$\pm$0.3 &  {73.9$\pm$0.4} & 96.3$\pm$0.7 &  {95.9$\pm$0.4} & 92.5$\pm$0.5 & {86.2$\pm$0.6 }& 96.8$\pm$0.6 &  {96.2$\pm$0.3} &  {88.4$\pm$0.5} &  {71.2$\pm$0.4 }&  {88.7$\pm$0.5} \\ 

{CAN + \textbf{SubUDA}+Dyn} ($m=5$) &  {97.4$\pm$0.5} &  {86.1$\pm$0.7} & 77.5$\pm$0.4 &  {74.2$\pm$0.6} & 96.2$\pm$0.3 &  {96.1$\pm$0.4} & 92.8$\pm$0.3 & {86.2$\pm$0.4 }& 96.6$\pm$0.5 &  {96.7$\pm$0.4} &  {89.0$\pm$0.6} &  {74.2$\pm$0.3 }&  {89.1$\pm$0.4} \\\hline\hline
 
SimNet (CVPR18) \cite{pinheiro2018unsupervised} & {94.3} & 82.3 & 73.5 & 47.2 & 87.9 & 49.2 & 75.1 & 79.7 & 85.3 & 68.5 & 81.1 & 50.3 & 72.9 \\

GTA (CVPR18) \cite{sankaranarayanan2018generate}& - & - & - & - & - & - & - & - & - & - & - & - & 77.1\\\hline

\textbf{SubUDA}+Dyn($m=5$): Res152 & {94.0$\pm$0.6} & {87.1$\pm$0.7} & {75.4$\pm$0.5}& {74.7$\pm$0.6 }& {90.8$\pm$0.6} & {89.5$\pm$0.7} & {88.8$\pm$0.8 }& {85.1$\pm$0.4 }& {91.8$\pm$0.6} & {93.5$\pm$0.8} &  {86.5$\pm$0.7} & {80.6$\pm$0.6} & {83.4$\pm$0.5 }\\

{\textbf{SubUDA}($m=5$): Res152 \cite{Liu2021subtype}} & {93.9$\pm$0.6} &  {86.8$\pm$0.7} & {75.2$\pm$0.5}&  {74.4$\pm$0.6 }& {90.4$\pm$0.6} &  {89.4$\pm$0.7} & {88.2$\pm$0.8 }& {85.0$\pm$0.4 }&  {91.4$\pm$0.6} &  {93.2$\pm$0.8} &  {86.1$\pm$0.7} &  {80.4$\pm$0.6} &  {83.1$\pm$0.5}\\

\textbf{SubUDA}-$K_n$ (validated $K_n$): Res152 & 93.3$\pm$0.5 & 85.6$\pm$0.6 & 74.5$\pm$0.5 & {74.7$\pm$0.8} & 90.6$\pm$0.9& {83.5$\pm$0.7} & {89.4$\pm$0.5} & {83.8$\pm$0.6} & 89.1$\pm$0.7 &  {92.9$\pm$0.5} & {86.6$\pm$0.8} & 74.5$\pm$0.5 & 83.2$\pm$0.5\\

\hline

\end{tabular}%
}
 
\caption{Experimental results for VisDA17-val setting. We use ResNet101 as our backbone except for \cite{pinheiro2018unsupervised,sankaranarayanan2018generate}, which uses Res152. Note that Res50 used in some methods is a stronger baseline than Res101. Our proposed SubUDA and its variants are bold.}
\label{table:visda17}
\end{table*}

\begin{table*}[!t]
\centering
\resizebox{0.8\linewidth}{!}{
\centering
\begin{tabular}{l|cccccccccccc}
\hline
Class        &  Aeroplane & Bicycle   & Bus & Car & Horse & Knife & Motor   & Person   & Plant & Skate Board & Train  & Truck    \\ \hline\hline
$K_n$ & 6 & 3 & 4 & 5 & 5 & 4 & 3 & 4 & 6 & 3 & 2 & 4  \\\hline
\end{tabular}%
}
\caption{Validated $K_n$ for each class in VisDA dataset with ResNet101 backbone.}
\label{table:visda17statistic}
\end{table*}

As shown in Table \ref{tabel:chd}, the SubUDA ($m=8$) adopted the reliability-path based online sub-graph as an alternative to $K$-means. With an appropriate $m$, e.g., $m=8$, the adaptively learned clustering achieved the performance on par with the $K$-means with $K_n=4$. We can see from Fig. \ref{fig:66} right that the semi-hard mining scheme in SubUDA ($m=8$) is not sensitive to $\tau$ for a relatively large range. This may be attributed by the fact that the network can flexibly learn to adjust the ratio of $\epsilon$ and $\tau$ in mapping space \cite{liu2017adaptive}. We note that the too strict semi-hard mining scheme, i.e., too small $\tau$, can degenerate our SubUDA to conventional class centroids matching, since no target samples are selected to form a subtype cluster.

A hyper-parameter sensitive analysis w.r.t. $K_n$ and $m$ for two kinds of online clustering schemes is provided in Fig. \ref{fig:77}. Consensus clustering is widely used to assess the clustering stability \cite{monti2003consensus}, and the optimal clustering is usually achieved in the elbow position of the area under CDF changes, where the CDF is for the consensus matrices\footnote{\url{https://github.com/ZigaSajovic/Consensus_Clustering}}. We can see that the accuracy peak typically coincides with the best metric of clustering consensus, which clearly indicates that the SubUDA performance relied on the good subtype clustering. The choice of $K_n=4$ also matched our prior knowledge of CHD patient subtype numbers.

{The sensitivity analysis of $\epsilon$ in CHD and VisDA datasets are shown in Fig. \ref{fig:abla}. We can see that the accuracy is not sensitive to a relatively large range of $\epsilon$, as expected.} 

Since there were clearly four subtypes in our data, concise $K$-means can be used as our solution. However, Fig. \ref{fig:77} right shows that the accuracy curve is robust for a relatively large range of $m$, which is promising for the hyperparameter tuning of the case without the prior information of subtype numbers. Specifically, when we do not know $K_n$, we propose to use a sub-graph for clustering with $\epsilon=1$, $\tau$, and $m$ to decide the cluster numbers. Therefore, we only needed two hyperparameters ($\tau$ and $m$) rather than $N$.

From Fig. \ref{fig:55}, we can clearly see that samples from the same subtype are closely distributed, but not clearly separated. We only enforced subtype compactness, but did not enforce subtype separation, since classification is our goal in this work.

\begin{figure}[t]
\centering
\includegraphics[width=9cm]{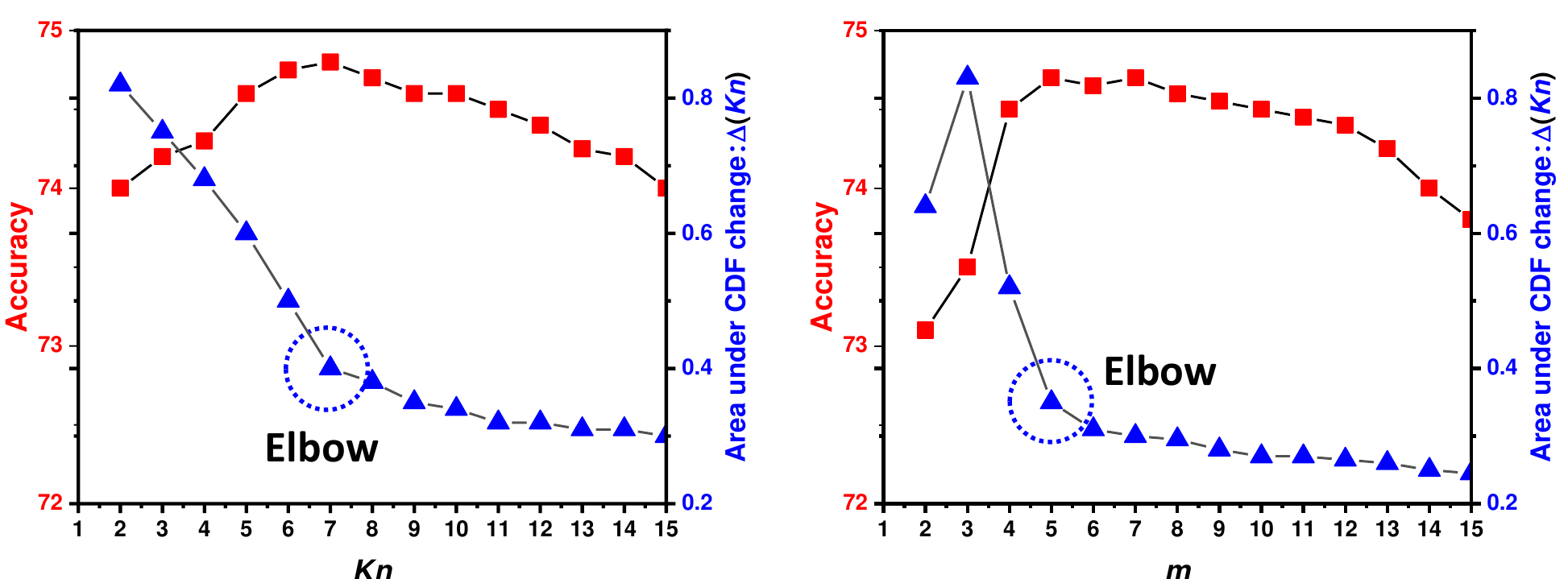}\\ 
\caption{Sensitive analysis of $K_n$ (right) and $m$ of car class in the VisDA17 dataset of dynamic SubUDA with the ResNet152 backbone. The red rectangle indicates accuracy and the blue triangle indicates AUC of CDF.}\label{fig:88}
\end{figure}

\subsection{Synthetic-to-Real Image UDA}

VisDA17 \cite{visda2017} is well-suited to evaluate our framework for a large scale case, where VisDA17 contains a synthetic source domain and a real image target domain. It contains a total of 280K images in 12 classes, in which a total of $152,409$ synthetic 2D images were used as the source domain and a total of $55,400$ real images were used as the target domain. This scale, however, poses a challenge to domain adaptation. Some examples are shown in Fig. \ref{fig:sp1}. For the car and truck classes, we can see that the inner-class variation is large, while the inter-class variation is relatively small. For fair comparisons with the other methods, we used the standard ResNet101 as the backbone, and set the batch size to 64 and $\mathcal{I}=5$. {We set $\alpha=1$, $\lambda=0.5$, and $\beta=0.5$ and used Adam with lr=1e-4.}

We did not know the subtype number in this task, and therefore the reliability-path-based sub-graph was utilized as our default solution. In fact, there is no clear definition of subtypes in VisDA, e.g., the car class can be grouped according to the maker, type, and color based on different taxonomy. Validating the best $K_n$ for all of the classes is tedious, and the optimal $K_n$ can be different when we change the backbone. 

In Fig. \ref{fig:88}, we investigate the effect of $K_n$ and $m$ on the car class. Given the ambiguous subtype grouping taxonomy, the accuracy curve w.r.t. $K_n$ is smoother than the CHD transfer task. Based on the accuracy and elbow position of consensus clustering analysis, we selected $m=5$ for our online sub-graph clustering. The performance was stable for a wide range of $m$, and the selected $m$ was shared for all subtypes and classes. Based on our experiments, the car class was more inclined to be clustered w.r.t. the vehicle type. In Table \ref{table:visda17statistic}, we provide the validated $K_n$ in the VisDA dataset. Since the two settings of $K_n$ had similar performance, we selected the smaller $K_n$ as the subtype number.

\begin{table*}[!t]
\centering
\resizebox{1\linewidth}{!}{
\centering
\begin{tabular}{l|cccccccccccc}
\hline
Divisions  &  Furniture & Mammal   & Tool & Cloth & Electricity & Building & Office & Human  & Road Trans & Food & Nature  & Cold Blooded    \\ \hline
Subtype & 35 & 25 & 28 & 23 & 25 & 21 & 21 & 20 & 15 & 14 & 14 & 12  \\\hline\hline
Divisions  &  Music & Fruit   & Sport & Tree & Bird & Vegetable & Shape & Kitchen  & Water Trans & Sky Trans & Insect  & Others  \\ \hline 
Subtype & 10 & 9 & 10 & 8 & 7 & 8 & 7 & 8 & 6 & 4 & 4 & 13  \\\hline
\end{tabular}%
}
\caption{Detailed statistics of each division in the DomainNet dataset. We regard the division and sub-class label as ``class" and ``subtype" label, respectively.}
\label{table:domainnet}
\end{table*}

\begin{table*}[!t]
\centering
\resizebox{1\linewidth}{!}{
\centering
\begin{tabular}{l|cccccccccccc}
\hline
Divisions  &  Furniture & Mammal   & Tool & Cloth & Electricity & Building & Office & Human  & Road Trans & Food & Nature  & Cold Blooded    \\ \hline
$K_n$ & 42 & 38 & 34 & 35 & 40 & 37 & 32 & 25 & 29 & 30 & 18 & 24  \\\hline\hline
Divisions  &  Music & Fruit   & Sport & Tree & Bird & Vegetable & Shape & Kitchen  & Water Trans & Sky Trans & Insect  & Others  \\ \hline 
$K_n$ & 20 & 28 & 13 & 18 & 12 & 14 & 16 & 10 & 22 & 12 & 15 & 38  \\\hline
\end{tabular}%
}
\caption{Validated $K_n$ in DomainNet dataset \cite{peng2019moment}.}
 
\label{table:domainnetvali}
\end{table*}

\begin{table*}[t]
\setlength{\tabcolsep}{0.085em}
\renewcommand{\arraystretch}{0.8}
{\resizebox{1\linewidth}{!}{
\centering

\begin{tabular}{c|c c c c c c c || c | c c c c c c c || c |c c c c c c c}
\hline
{TPN$^{meta}$} & {clp} & {inf} & {pnt} & {qdr} & {rel} & {skt} & {Avg.} &
{PANDA$^{meta}$} & {clp} & {inf} & {pnt} & {qdr} & {rel} & {skt} & {Avg.} &
{DMRL$^{meta}$} & {clp} & {inf} & {pnt} & {qdr} & {rel} & {skt} & {Avg.}  \\
\hline

{clp} & N/A	& 56.2 &	62.4 &	50.5 &	54.1 &	52.8 &	{56.4} & 
{clp} & N/A    & 58.2 &	64.4 &	51.2 &	55.8 &	57.6 &	{55.3}&
{clp} & N/A    & 57.7 &	65.4 &	52.3 &	58.2&	60.5&	{57.2}\\

{inf}&	55.8&	N/A&    48.5&	46.2&	67.4&	54.6&	{51.7}&	
{inf}&	57.2&	N/A&	45.5&	44.2&	63.1&	52.6&	{51.3}&	
{inf}&	57.5&	N/A&	48.9  &	46.7&   63.1&	54.3&	{52.7}\\

{pnt}&	66.1&	61.6&	N/A&	45.4&	72.6&	63.2&	{65.2}&  	
{pnt}&	69.8&	64.5&	N/A&	47.6&   75.1&	66.2&	{62.2}&	
{pnt}&	68.5&	66.3&   N/A&	50.5&   71.2&	67.2&	{63.0} \\

{qdr}&	42.2&	71.3&	62.5&	N/A&	47.5& 50.2&	{61.8}&
{qdr}&	42.6&	78.6&	65.2&	N/A&	50.6&52.0&	{61.7}&	
{qdr}&	41.0&	76.8	&68.4&	N/A&	50.1&	51.2&	{62.3}\\

{rel}&	60.5&	53.1&	64.5&	42.5&	N/A&	52.6&	{52.3}&	
{rel}&	61.4&	54.2&	65.4&	48.2&	N/A&	56.0&	{55.1}&	
{rel}&	61.5&	56.2&	66.5&	49.5	&N/A&	57.8&	{56.9}\\

{skt}&	53.6&	56.8&	48.4&	50.3&	64.6&	N/A&	{53.4}&	
{skt}&	52.0&	56.9	&46.2&	52.8	&64.1&	N/A&	{55.2}	&
{skt} &52.2&	58.3	&48.4&	55.3&	66.8&	N/A&	{58.1}\\

{All-Avg.}&	- &	- &	- &	- &	- &		- &	\underline{55.7}&
{All-Avg.}&	- &	- &	- &	- &	- &		- &	\underline{57.7}&
{All-Avg.}&	- &	- &	- &	- &	- &		- &	\underline{57.6}\\
\hline\hline

{\textbf{SubUDA}+Dyn$^{meta}_{m=10}$} & {clp} & {inf} & {pnt} & {qdr} & {rel} & {skt} & {Avg.} &
{\textbf{SubUDA}+Dyn$^{meta}_{m=10}$-$\omega_k$} & {clp} & {inf} & {pnt} & {qdr} & {rel} & {skt} & {Avg.} &
{\textbf{SubUDA}+Dyn$^{meta}_{m=10}$-$\tau$} & {clp} & {inf} & {pnt} & {qdr} & {rel} & {skt} & {Avg.}\\
\hline
{clp} & N/A	&    66.2 &	71.0 &	58.6 &	57.5 &	65.2 &	{62.1} & 
{clp} & N/A    & 66.6 &	69.2  &	57.3 &	56.9 &	65.6 &	{61.3}&
{clp} & N/A    & 65.3&	68.4 &	55.6 &	60.2&	66.7&	{62.0}\\

{inf}&	55.6&	N/A&	56.3&	43.6&	65.7&	52.4&	{56.6}&	
{inf}&	55.1&	N/A&	55.4&	43.2&	64.9&	51.6&	{56.0}&	
{inf}&	54.6&	N/A&	55.2&	43.5&   64.7&	52.0&	{56.2}\\

{pnt}&	62.4&	65.6&	N/A&	51.6&	78.2&	62.6& 	{67.2}&  	
{pnt}&	61.2&	69.5&	N/A&	48.4&    79.1&	65.4&	{66.7}&
{pnt}&	61.4&	64.8	& N/A	&50.9&	78.5&	62.4&	{66.0} \\

{qdr}&	44.1&	78.8&	66.4&	N/A&	55.6& 58.6&	{61.8}&
{qdr}&	43.7&	78.6&	65.4&	N/A&	55.5&  58.6&	{61.1}&	
{qdr}&	44.0&	78.0	&66.0&	N/A&	55.5&	58.2&	{61.3}\\

{rel}&	67.3&	52.6&	69.5&	57.5&	N/A&	52.8&	{55.8}&	
{rel}&	66.5&	51.5&	69.1&	57.1&	N/A&	51.7&	{54.2}&	
{rel}&	66.6&	51.6&	68.6&	56.2	&N/A&	52.3&	{54.2} \\

{skt}&	52.3&	61.6&	58.2&	52.6&	70.3&	N/A&	{60.8}&	
{skt}&	52.3&	60.9	& 58.2&	51.9	&70.6&	N/A&	{60.4}	&
{skt} &51.2&	61.7	&57.6&	51.8&	70.8&	N/A&	{60.2} \\

{All-Avg.}&	- &	- &	- &	- &	- &		- &	\underline{63.9}&
{All-Avg.}&	- &	- &	- &	- &	- &		- &	\underline{61.8}&
{All-Avg.}&	- &	- &	- &	- &	- &		- &	\underline{62.2} \\\hline\hline

{\textbf{SubUDA}+Dyn$^{meta}_{K_n}$} & {clp} & {inf} & {pnt} & {qdr} & {rel} & {skt} & {Avg.} &
{\textbf{SubUDA}+Dyn$^{meta}_{Vali-K_n}$} & {clp} & {inf} & {pnt} & {qdr} & {rel} & {skt} & {Avg.} &
{TPN$^{sub}$} & {clp} & {inf} & {pnt} & {qdr} & {rel} & {skt} & {Avg.}\\
\hline
{clp} & N/A	&    67.7 &	72.5 &	59.8 &	60.7 &	66.5 &	{63.4} & 
{clp} & N/A    & 66.6 &	70.2  &	59.3 &	62.4 &	69.6 &	{64.3}&
{clp} & N/A    & 68.3&	71.4 &	58.6 &	63.2&	69.7&	\textbf{65.0}\\

{inf}&	57.4&	N/A&	58.3&	45.2&	67.5&	54.2&	{58.2}&	
{inf}&	58.1&	N/A&	58.5&	45.3&	67.9&	54.2&	{58.5}&	
{inf}&	59.5&	N/A&	59.7&	49.6&   68.8&	58.5&	\textbf{59.4}\\

{pnt}&	62.4&	67.6&	N/A&	55.6&	79.2&	64.6& 	{68.5}&  	
{pnt}&	61.2&	60.5&	N/A&	51.4&    79.1&	65.4&	\textbf{68.6}&
{pnt}&	60.2&	62.3	& N/A	&52.5&	77.0&	64.1&	{67.1} \\

{qdr}&	51.2&	79.1&	67.8&	N/A&	58.4&   60.2&	{63.5}&
{qdr}&	51.5&	79.1&	67.5&	N/A&	58.6&   60.5&	\textbf{63.7}&	
{qdr}&	50.8&	79.2	&67.9&	N/A&	56.8&	59.7&	{62.3}\\

{rel}&	69.4&	55.5&	70.2&	59.4&	N/A&	54.5&	{57.5}&	
{rel}&	69.6&	56.4&	70.3&	59.2&	N/A&	54.6&	{57.6}&	
{rel}&	70.5&	55.7&	71.3&	60.4	&N/A&	56.2&	\textbf{58.3} \\

{skt}&	53.5&	63.8&	58.3&	53.3&	71.4&	N/A&	{61.4}&	
{skt}&	53.4&	64.7	& 58.2&	53.8	&71.7&	N/A&	{61.7}	&
{skt} &52.6&	62.4	&58.5&	52.9&	71.6&	N/A&	\textbf{61.8} \\

{All-Avg.}&	- &	- &	- &	- &	- &		- &	\underline{64.2}&
{All-Avg.}&	- &	- &	- &	- &	- &		- &	\underline{64.5}&
{All-Avg.}&	- &	- &	- &	- &	- &		- &	\underline{64.9} \\\hline\hline

\end{tabular}
}}
\caption{Single-source meta-class evaluations on the DomainNet \cite{peng2019moment}. Several single-source adaptation baselines are validated on the DomainNet-metaclass dataset, including TPN \cite{pan2019transferrable}, PANDA \cite{hu2020panda}, and DMRL \cite{wu2020dual}. The TPN$^{sub}$ utilized the subtype label as a supervision signal, which can be regarded as an upper bound. The left column denotes the source domain, and the top row denotes the target domain, respectively. The Ave. in the most right column represents the average accuracy of each row. The All-Ave. in the bottom denotes the average performance of all domain combinations.}
\label{tab_challengeI}
\end{table*}

Table \ref{table:visda17} tabulates the accuracy using VisDA17. We used suffixes +Dyn, -$\omega_k$, and -$\tau$ to denote SubUDA with queue memory, without subtype balance weight, and without semi-hard target mining, respectively. Besides, the suffix -$\mu_k^{st}$ denotes using $\mu_k^{st}=\frac{\sum_{i=1}^{M_k^s+M_k^t}f(x_i^{st})}{M_k^s+M_k^t}$ as the subtype centroid, which was not robust against the subtype label shift. For each proposed approach, we independently ran the test five times, and the average and standard deviation (SD) of the accuracy metrics were reported. 

All SubUDAs, including the $K$-means and sub-graph based methods, outperformed TPN \cite{pan2019transferrable} by around 2\% w.r.t. mean accuracy, demonstrating the effectiveness of subtype-wise compactness. SubUDA achieved even superior performance over the methods with the stronger ResNet-152 backbone \cite{pinheiro2018unsupervised,sankaranarayanan2018generate}. We also adopted ResNet-152 as our backbone to demonstrate the generalizability of SubUDA. The improvement was consistent with the ResNet101 backbone. {CAN \cite{kang2019contrastive} has been an powerful baseline for image classification UDA. Notably, CAN introduced the inter-class domain discrepancy measure which improves the mean accuracy in VisDA-17 over the CAN with only inter-class term by 3.3\%. In addition, there are several training tricks used in CAN to boost the performance, e.g., alternative optimization and class-aware sampling. We also used CAN as our baseline and added the subtype compactness objectives to further boost the performance by 1.9\%.}

Similar to the CHD transfer task, the dynamic queue scheme substantially boosted adaptation performance. Without a clear subtype definition, the semi-hard target mining may also be helpful to reject confusing samples. Considering that there can be many underlying subtypes, balancing the cluster size with $\omega_k$ is also necessary. Besides, as the more subtypes lead to fewer samples in each subtype cluster, the more severe the subtype label shift is likely to occur. Therefore, calculating $\mu_k^{st}$ in a subtype label shift robust manner is crucial. We did not test SubUDA-DR, since its clustering speed was not scalable to large-scale datasets.

Converting clustering into deep clustering \cite{caron2018deep} is highly challenging and costly. With our proposed dynamic online clustering and dimension reduction scheme, training of VisDA took about 10 hours, while CRST \cite{zou2019confidence} took about 8 hours. We note that the speed of testing was the same as conventional alternatives. The memory modules used in the dynamic online scheme only stored the extracted feature vector, which required 1.4$\times$ more memory than vanilla TPN. With an acceptable and practical training cost, the improvement was substantial.

\begin{table*}[!t]
\centering
\caption{{Experimental results for GTA5 to Cityscapes.}}
\resizebox{1\linewidth}{!}{
\centering
\begin{tabular}{c|ccccccccccccccccccc|c}
\hline
Method                  & Road & SW   & Build & Wall & Fence & Pole & TL   & TS   & Veg. & Terrain & Sky  & PR   & Rider & Car  & Truck & Bus  & Train & Motor & Bike & mIoU \\ \hline\hline

Source  & 81.8& 16.3& 74.4 &18.6& 12.7& 23.5 &29.3 &18.1 &73.5 &21.4 &77.6 &55.6 &25.6 &74.1 &28.6 &10.2 &3.0 &25.8 &32.7 &37.0\\ \hline

SAPNet\cite{li2020spatial}  & 88.4 &38.7& 79.5& 29.4& 24.7& 27.3& 32.6& 20.4& 82.2& 32.9& 73.3& 55.5& 26.9& 82.4& 31.8 &41.8 &2.4 &26.5& 24.1& 43.2\\ \hline

OCE\cite{cordts2016cityscapes} & 89.4& 30.7 &82.1 &23.0 &22.0 &29.2 &37.6 &31.7 &83.9 &37.9 &78.3 &60.7 &27.4 &84.6 &37.6 &44.7 &7.3 &26.0 &38.9 &45.9\\ \hline

\textbf{SubUDA}+Dyn & 93.1 & 44.9 & 81.6 & 32.6 & 26.5 & 31.4 & 38.0 & 32.8 & {87.6} & 36.9 & 79.4 &{64.3}& 32.4 & 85.0 & 23.5 & 43.6 & 15.2 & 32.2 & 39.4 & 49.6 \\\hline\hline

\end{tabular}%
}

\label{table:gtacity}
\end{table*}

\subsection{Large scale two-level UDA in DomainNet}

DomainNet \cite{peng2019moment} is the largest domain adaptation dataset to date, which consists of $\sim$0.6M images from 24 divisions (meta-class). There are the second-level labels of 345 sub-classes for 24 meta-class as listed in Table \ref{table:domainnet}. DomainNet has 6 distinct domains, i.e., \text{Clipart} ({{clp}}), \text{Infograph} ({{inf}}), \text{Painting} ({{pnt}}), \text{Quickdraw} ({{qdr}})\footnote{\url{https://quickdraw.withgoogle.com/data}}, \text{Real} ({{rel}}), and \text{Sketch} ({{skt}}). For each domain, the standard evaluation protocol split the dataset, following a 70\%/30\% scheme for training and testing sets. The number of instances for both 24 meta-classes and 345 sub-classes are different\footnote{\url{http://ai.bu.edu/M3SDA/}}, which introduces the label shifts at both class and subtype levels. 

To utilize the two-level label system in DomainNet to investigate the subtype-aware UDA, we combined the sub-classes with corresponding meta-classes in a division into corresponding meta-classes and assumed that the sub-classes are not accessible. Therefore, we only have the meta-class label in the source domain, and the classifier works for meta-class prediction. The sub-class within a meta-class can be greatly different from one another. For example, sport meta-class includes baseball, yoga, hockey puck, etc. It is difficult to extract the shared pattern across these sub-classes. With this setting, it is possible to investigate the effect of subtype on large-scale datasets.

We re-implemented several recent methods for the meta-class classification, including TPN \cite{pan2019transferrable}, PANDA \cite{hu2020panda}, and DMRL \cite{wu2020dual}. The superscript ``meta" means we only used meta class label, and the subtype label was not accessible. {We set $\alpha=1$, $\lambda=0.5$, and $\beta=0.5$ and used Adam with lr=1e-4.}

\begin{figure}[t]
\centering
\includegraphics[width=8cm]{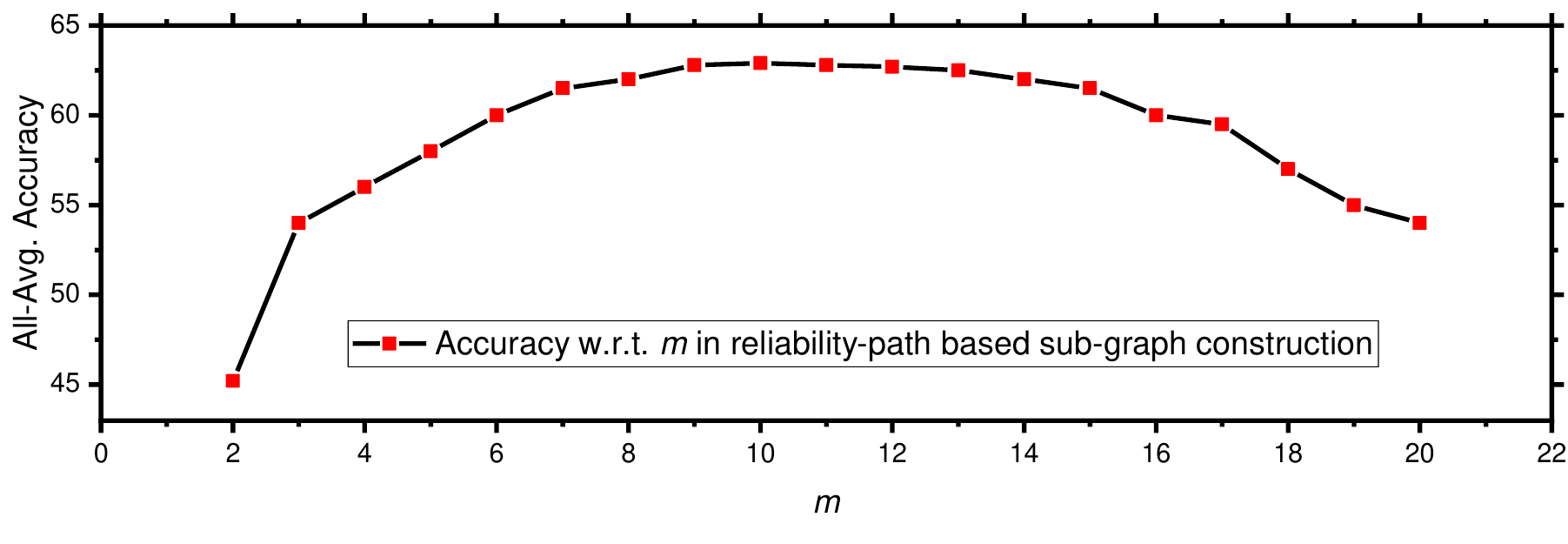}\\ 
\caption{Sensitive analysis of $m$ in DomainNet with the ResNet101 backbone.}\label{fig:domainnet}
\end{figure}

To compare our framework with the subtype label supervision method, we used sub-class and grouped the prediction of sub-classes as the meta-class prediction. We note that the subtype label supervised setting can empirically be an ``upper-bound" of our subtype label unsupervised setting. We denote this setting with the superscript ``sub."

We used suffixes -$\omega_k$ and -$\tau$ to denote the subtype balance weight and semi-hard target mining, respectively. We note that we cannot efficiently or practically implement the clustering in DomainNet without either the dynamic memory scheme or dimension reduction. We also assumed that $K_n$ was unknown, and thus used the adaptive sub-graph construction. We validated the choice of $m$ as shown in Fig. \ref{fig:domainnet}, and consistently chose $m=10$.

By explicitly considering the underlying subtypes, our SubUDA+Dyn outperformed the baseline TPN \cite{pan2019transferrable} and the recent SOTA methods \cite{hu2020panda,wu2020dual}. Since we knew the number of sub-class in DomainNet, we simply assigned $K_n$ in $K$-means, according to the statistics in Table \ref{table:domainnet}, which is denoted by SubUDA+Dyn$_{K_n}^{meta}$. With the prior knowledge of subtype number, the performance is better than the adaptive sub-graph construction based SubUDA+Dyn$_{m=10}^{meta}$. The higher accuracy of SubUDA+Dyn$_{m=10}^{meta}$ over SubUDA+Dyn$_{m=10}^{meta}$-$\omega_k$ and SubUDA+Dyn$_{m=10}^{meta}$-$\tau$ clearly indicates the effectiveness of subtype balance weight and the semi-hard target mining.

The variation within a sub-class in DomainNet can be large. For instance, the sub-class of the airplane can involve jet, turboprop, single-engine aerobatic, etc. In Table \ref{table:domainnetvali}, we provide the validated $K_n$ in the VisDA dataset. For the two settings where $K_n$ had similar performance, we selected the smaller $K_n$ as the subtype number. We can find that the validated $K_n$ is consistently larger than the subclass number, which indicates some of the subclasses can have large variations and should be further split. 

With the validated $K_n$, SubUDA+Dyn$_{Vali-K_n}^{meta}$ achieved a state-of-the-art performance. More appealingly, it outperformed the subtype label supervised TPN$^{sub}$. We note that fine-grained labeling is costly and noisy in real-world applications. Moreover, validating $K_n$ is costly in large-scale datasets. Considering the relatively small improvement of SubUDA+Dyn$_{Vali-K_n}^{meta}$ over SubUDA+Dyn$_{K_n}^{meta}$, we may not need to know the number of sub-class.

\begin{figure}[t]
\centering
\includegraphics[width=6cm]{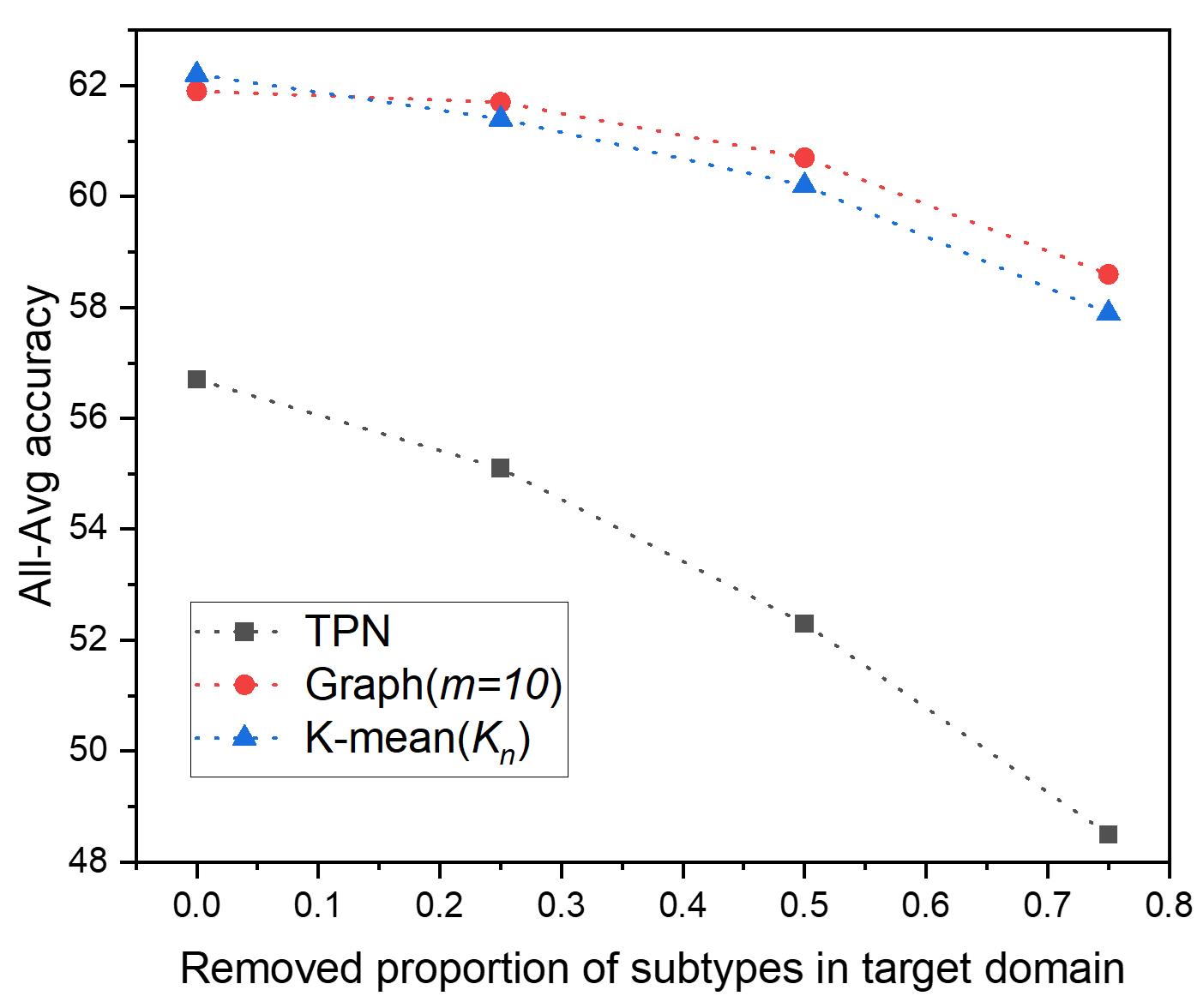}\\ 
\caption{Sensitive analysis of subtype difference in source and target domains of TPN (gray) \cite{pan2019transferrable}, SubUDA+Dyn$_{m=10}^{meta}$ (red), and SubUDA+Dyn$_{K_n}^{meta}$ (blue).}\label{fig:sensi}
\end{figure}

{Our SubUDA framework does not require that the classes or subtypes are the same in the source and target domains. To investigate the robustness of our framework in the case where only a subset of subtypes exist in the target domain, we investigated the performance of removing 0\%, 25\%, 50\%, and 75\% subtypes in the training set of the target domain. We used the shared 25\% subtypes in each class in our testing set at the testing stage. The subtypes in testing existed in all of training settings. The result is shown in Fig.~ \ref{fig:sensi}. We can see that the performance of both TPN and our SubUDA drops due to the label shift \cite{liu2021adversarial}. However, our subtype clustering framework was less affected by the label shift, compared with TPN \cite{pan2019transferrable}.}

{Though our method does not require the number of categories to be the same between two domains, it is not specifically designed for the partial UDA. We are exploring the subtype, which is a more fine-grained level of the category. Therefore, it is more reasonable to investigate the label shift of subtype rather than category (e.g., partial UDA), which is evidenced in Fig.~ \ref{fig:sensi}. In addition, for partial UDA tasks \cite{liu2021adversarial}, the performance of our SubUDA is similar to the TPN.}

\subsection{{UDA for segmentation}}

{We then investigated the effectiveness of the subtype compactness in segmentation, which is equivalent to pixel-wise classification. We evaluated our framework on the GTA5 \cite{richter2016playing} to Cityscapes \cite{cordts2016cityscapes} task, which has 19 shared segmentation classes. There are more than 24,000 labeled game engine rendered images in the GTA5 dataset. We followed the evaluation protocol and backbones as in \cite{toldo2021unsupervised}, and adopted the cropping strategy in ROAD \cite{chen2018road} to enlarge the batch size to 10 as in \cite{chen2018road}. We note that representing each pixel with a feature vector and clustering them can be quite costly. For example, the batch size in \cite{cordts2016cityscapes} is set to 1. We simply set $m=20$ for all classes in our SubUDA+Dyn. We note that, similar to DomainNet, SubUDA without Dyn was not scalable to the segmentation task.}

{We compare our SubUDA+Dyn with the other methods in Table \ref{table:gtacity}. Based on the class-wise compactness method of \cite{cordts2016cityscapes}, SubUDA outperformed \cite{cordts2016cityscapes}, by 3.7\% w.r.t. the mean IoUs. In comparison, \cite{cordts2016cityscapes} was focused on the class-wise compactness, which is different from our subtype-wise structure. The point-wise feature similarity within a class can be problematic, e.g., simply enforcing the similarity between the spatial feature related to car's window and car's wheel. Actually, in \cite{cordts2016cityscapes}, the samples in each cluster were known, and we only needed to calculate the distance. We note that this is the case when we have the label in the source domain and perform supervised clustering. However, it would be more challenging for unsupervised clustering (e.g., sub-graph) to identify the examples in each cluster without the subtype label in both domains.}

\section{Conclusions}\label{sec:conc}

In this work, we presented a novel approach to adaptively carry out the fine-grained subtype-aware alignment in UDA, based on the observations that subtype-wise conditional and label shifts widely exist, and those shifts can be adaptively aligned without the subtype label. We proposed a flexible yet principled approach for the case with or without the prior knowledge on the number of subtypes. In addition to concise $K$-means, we further extended our approach with an online sub-graph scheme using a reliability-path, which can be scalable to many classes and subtypes with only a few meta hyperparameters. We also demonstrated that our alternatively updated dynamic memory queue and centroids modules could effectively process the imbalanced and undersampled subtypes. Our thorough evaluations on CHD transfer and large-scale general UDA tasks clearly demonstrated its validity and superiority to the recent SOTA UDA approaches.

\ifCLASSOPTIONcaptionsoff
  \newpage
\fi

\bibliographystyle{IEEEtran}
\bibliography{main}

\end{document}